\documentclass{article}

\PassOptionsToPackage{numbers, compress}{natbib}
    \usepackage[preprint]{neurips_2025}

\usepackage[utf8]{inputenc} % allow utf-8 input
\usepackage[T1]{fontenc}    % use 8-bit T1 fonts
\usepackage[hidelinks]{hyperref}       % hyperlinks
\usepackage{url}            % simple URL typesetting
\usepackage{booktabs}       % professional-quality tables
\usepackage{amsfonts}       % blackboard math symbols
\usepackage{nicefrac}       % compact symbols for 1/2, etc.
\usepackage{microtype}      % microtypography
\usepackage{xcolor}         % colors
\usepackage{amsthm}         % theorems
\usepackage{amsmath}
\usepackage{multirow}
\usepackage{arydshln}
\usepackage{graphicx}
\usepackage{placeins}  % in preamble
\usepackage{subcaption}
\usepackage{amssymb}
\usepackage{mathtools}
\usepackage[ruled,vlined,linesnumbered]{algorithm2e}

\SetKwInput{KwRequire}{Require}
\SetKwInput{KwReturn}{Return}
\SetAlgoNlRelativeSize{-1}

\theoremstyle{plain}
\newtheorem{theorem}{Theorem}[section]
\newtheorem{proposition}[theorem]{Proposition}

\title{Exploiting the Exact Denoising Posterior Score in Training-Free Guidance of Diffusion Models}

\author{%
  Gregory D. Bellchambers \\
  PhysicsX Ltd.\\
  London, EC2A 4DQ \\
  \texttt{greg.bellchambers@physicsx.ai} \\
}

\begin{document}

\maketitle

\begin{abstract}
% Similarly, a novel formulation of the posterior score for noisy inpainting tasks is derived in terms of an unconditional non-isotropic score function

The success of diffusion models has driven interest in performing conditional sampling via training-free guidance of the denoising process to solve image restoration and other inverse problems. A popular class of methods, based on Diffusion Posterior Sampling (DPS), attempts to approximate the intractable posterior score function directly. In this work, we present a novel expression for the exact posterior score for purely denoising tasks that is tractable in terms of the unconditional score function. We leverage this result to analyze the time-dependent error in the DPS score for denoising tasks and compute step sizes on the fly to minimize the error at each time step. We demonstrate that these step sizes are transferable to related inverse problems such as colorization, random inpainting, and super resolution. Despite its simplicity, this approach is competitive with state-of-the-art techniques and enables sampling with fewer time steps than DPS.
\end{abstract}

\section{Introduction}

Diffusion Models (DMs) \citep{song2019generative, ho2020denoising, karras2022elucidating} are a state-of-the-art class of generative model, achieving high quality, diverse sampling of complex data distributions. A particularly successful application is in conditional generation of image data, enabling rapid progress in class-conditioned image generation \citep{dhariwal2021diffusion,ho2022cascaded,zhang2023adding}, image editing \citep{huang2024diffusion} and image restoration tasks \citep{saharia2022image,song2020score,DPS}. Training-free guidance methods \citep{DPS, mcg,ddrm} attempt to avoid the high cost of problem-specific model training by steering the reverse diffusion process towards conditional samples. The exact guidance vector required to sample from the conditional distribution is the noisy likelihood score function, obtained by applying Bayes’ rule to the posterior score of the conditional distribution\citep{dhariwal2021diffusion}. The main challenge for training-free guidance is the intractability\citep{gupta2024diffusion} of the noisy likelihood score function.

In this paper, we focus on the approach of directly approximating the noisy likelihood score with application to inverse problems in the field of image restoration. The image degradation model involves application of a measurement operator followed by addition of Gaussian noise, and so the likelihood for clean data is given by a Normal distribution. At each time step of reverse diffusion, the DM unconditional score is augmented with a guidance term corresponding to the approximate likelihood score. When the measurement operator is simply the identity, the inverse problem is image denoising. By exploiting the structure of the noise-perturbed posterior score function, we show that the exact score for denoising is tractable in terms of the unconditional score function at all time steps. With access to the denoising posterior score, we can compute the exact noisy likelihood score for denoising tasks and evaluate the accuracy of existing methods on this task, as well as improve such methods in related tasks. To demonstrate the value of the tractable denoising score, we develop a method, DPS-w, for correcting DPS \citep{DPS} for tasks with significant denoising character, such as colorization, inpainting and super resolution. We hope that the results presented herein can inform future developments of principled training-free guidance methods.

The main contributions of this paper are:
\begin{enumerate}
\item{
A novel expression for the tractable denoising posterior score in terms of the unconditional DM score. We also present the result for inpainting in terms of the score function for a non-isotropic noising process \citep{gerdes2024gud}. Several exact conditions of the intractable score for inpainting are presented.
}
\item{
We use the norm of the exact posterior score to assess the step size heuristics adopted by DPS and other methods, showing that they result in guidance steps far larger than those implied by the true score for the majority of time steps.
}
\item{
We develop a simple method, DPS-w, to highlight the informative value of the tractable posterior score. For a reference denoising task, DPS step sizes are fit to the exact posterior score at each time step and transferred to related inverse linear problems. Despite its simplicity and lack of fine-tuned parameters, DPS-w is competitive with start-of-the-art methods on random inpainting and super resolution tasks. It is shown to be robust across a range of measurement noise levels, have little computation overhead, and enable sampling with a reduced number of steps.
}
\end{enumerate}

\section{Background}
\subsection{Diffusion models}
Diffusion Models (DMs) \citep{song2019generative, ho2020denoising, karras2022elucidating} involve a predefined Gaussian noising process that incrementally maps clean data $x_0$ at time $t=0$ to pure, isotropic noise $x_T \sim \mathcal{N}(0, I)$ at time $T$. To generate samples, the process is run in reverse, starting from sampled $x_T$, removing the noise $\epsilon_t(x_t)$ at each time step until a clean sample $x_0$ is obtained. A deep learning model is trained to predict the noise $\epsilon_\theta(x_t, t) \approx \epsilon_t(x_t)$.

In DDPM \citep{ddpm}, a Variance-Preserving (VP) process is adopted and the noised data has posterior distribution $q_t(x_t | x_0) = \mathcal{N}(x_t; \sqrt{\bar{\alpha}_t}x_0, (1 - \bar{\alpha}_t)I)$, where $\bar{\alpha}_t$ is a parameter of the noise schedule. The marginal noise-perturbed distribution is therefore
\begin{equation}
    p_t(x_t) = \int p(x_0) \mathcal{N}(x_t, \sqrt{\bar{\alpha}_t}x_0, (1 - \bar{\alpha}_t)I) dx_0,
\end{equation}
with $p(x_0)$ the target distribution of clean data $x_0$ \citep{song2020score}. Other methods \citep{song2019generative, song2020improved} adopt a Variance-Exploding (VE) process, with noise-perturbed distribution $p_t(x_t) = \int p(x_0)\mathcal{N}(x_t; x_0, \sigma^2_t I) dx_0$. The variance $\sigma^2_t$ is the noise level at time $t$. The distributions for VP and VE have been shown to be equivalent \citep{ddrm}.

The noise at time $t$ is related to the score of $p_t(x_t)$. In the case of DDPM, the score is given by \citep{dmreview}:
\begin{equation}
    \nabla_{x_t}\log p_t(x_t) = - \frac{1}{\sqrt{1-\bar{\alpha}_t}}\epsilon_t.
\end{equation}
Therefore, knowledge of the noise-perturbed score function is sufficient to generate samples via the denoising process. Similarly, we can define the approximate score function $s_\theta(x_t, t) = - \epsilon_\theta(x_t, t) / \sqrt{1-\bar{\alpha}_t}$.

\begin{figure}[t]
\label{fig:restorations}
\centering
\includegraphics[width=0.8\linewidth]{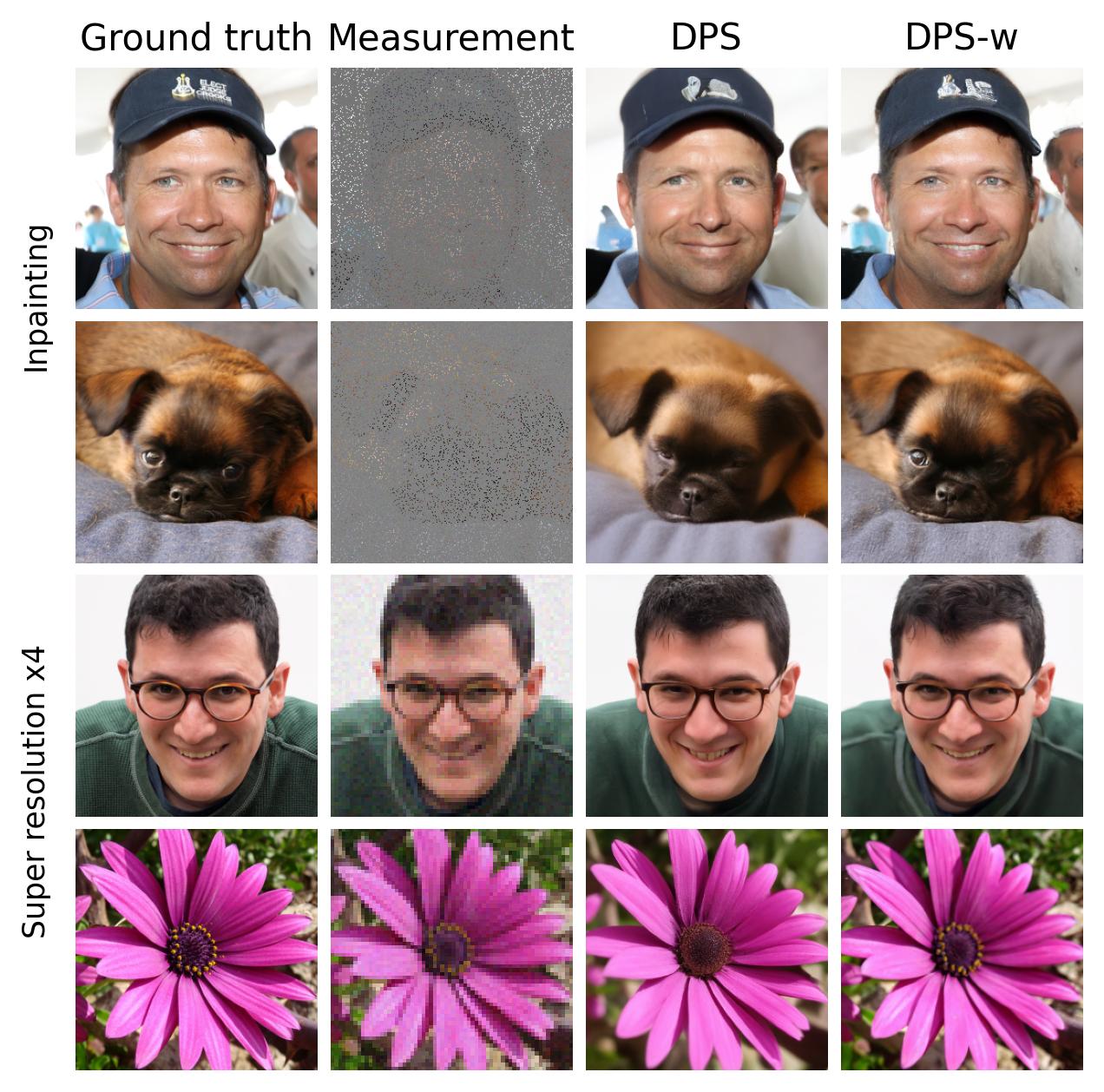}
\caption{Selected samples for inverse linear problems inpainting and super resolution with methods DPS and DPS-w, $\sigma_y = 0.05$.}
\end{figure}

\subsection{Training-free conditional diffusion models for inverse problems}
For a noisy measurement $y$ with forward model
\begin{equation}
\label{eqn:forward_model}
    y = \mathcal{A}(x_0) + \sigma_y \eta,
\end{equation}
where $\mathcal{A}$ is the measurement operator, $\sigma_y$ is the measurement noise level and $\eta \sim \mathcal{N}(0, I)$, we aim to solve the inverse problem to find realistic solutions $x_0$. Expressed in a Bayesian formulation, we aim to sample from the posterior $p(x_0 | y) \propto p(x_0) p(y | x_0)$, given the prior $p(x_0)$ and the likelihood $p(y|x_0) = \mathcal{N}(y; \mathcal{A}(x_0), \sigma^2_y I)$. To sample from the posterior with a reverse diffusion process, we need the noise-perturbed posterior score function, which is related to the noise-perturbed prior and likelihood score functions by \citep{dhariwal2021diffusion}
\begin{equation}
\label{eqn:score_bayes}
    \nabla_{x_t}\log p_t(x_t | y) = \nabla_{x_t}\log p_t(x_t) + \nabla_{x_t}\log p_t(y | x_t).
\end{equation}
Eq (\ref{eqn:score_bayes}) provides a route for sampling the posterior: at each time step, augment the unconditional DM score function with the guidance of the noisy likelihood score. In general, the likelihood score is intractable \citep{DPS, gupta2024diffusion} and needs to be approximated \citep{daras2024survey}.

\textbf{Diffusion Posterior Sampling (DPS)} \citep{DPS} introduces a popular approximation approach. First, it is observed that the time-dependent likelihood
\begin{equation}
\label{eqn:noisy_likelihood}
    p_t(y|x_t) = \int p(x_0|x_t)\,p(y|x_0) \, dx_0
\end{equation}
can be interpreted as the expectation $E_{x_0 \sim p(x_0|x_t)}[p(y|x_0)]$. Second, the expectation of the function $p(y|x_0)$ is approximated as the function of the expectation: $p(y|\hat{x}_0)$, where $\hat{x}_0 = E_{x_0\sim p(x_0|x_t)}[x_0]$. The posterior mean $\hat{x}_0$ is the MMSE estimation of $x_0$ given $x_t$, and is calculated in terms of the score function via Tweedie's formula \citep{kim2021noise2score}. For DDPM, $\hat{x}_0 = \frac{1}{\sqrt{\bar{\alpha}}_t}(x_t + (1 - \bar{\alpha}_t) \nabla_{x_t} \log p_t (x_t))$. The source of error in the DPS approximation is due to the Jensen gap.

The practical form of the DPS approximation given a pre-trained DM $s_\theta$ is
\begin{equation}
\label{eqn:dps_prac}
    \nabla_{x_t}\log p_t(x_t | y) \approx s_\theta(x_t, t) - \rho \nabla_{x_t} \| y - \mathcal{A}(\hat{x}_0) \|^2,
\end{equation}
and the step size $\rho = 1/\sigma^2_y$ is generally replaced with the time-dependent step size $\zeta_t = \zeta^\prime / \| y - \mathcal{A}(\hat{x}_0) \|$. Since $\hat{x}_0$ is a function of $x_t$, backpropagation through the neural network is required.

Finally, though DPS is applied to both linear and nonlinear inverse problems, we only consider linear operators $\mathcal{A}(x_0) = Ax_0$ in the rest of this paper.

\section{Related work}
There is a growing body of work on approximations to the noisy likelihood score, Eq~(\ref{eqn:noisy_likelihood}) \citep{daras2024survey}. MCG \citep{mcg} introduced the use of Tweedie's formaula to evaluate the likelihood for posterior mean, $\hat{x}_0$, which was extended to general noisy inverse problems by DPS \citep{DPS}. LDG \citep{LGD} draws Monte Carlo samples from a Normal distribution to reduce the bias in DPS, and extends application to general loss-function guidance. PGDM \citep{song2023pseudoinverse} modifies the Jensen approximation DPS to model $p(x_0|x_t)$ as a Gaussian, and introduces time-dependent variances, or step sizes, based on a heuristic. DSG \citep{DSG} introduces the concept of manifold deviation during the sampling process and proposes a constraint to keep the guidance step within a zone of high-confidence. DAPS \citep{DAPS} considers an alternative, noise-annealing process that decouples consecutive time steps, allowing large variation between steps. DAPS and DSG are recent state-of-the-art methods that are shown to out perform a wide range of established methods for the benchmark tasks considered in this paper.

\section{Method}
We present novel expressions for the posterior score function, Eq (\ref{eqn:score_bayes}), for pure denoising and noisy inpainting problems. The denoising posterior score is tractable when we have access to the score function for the isotropic noising process of DMs. The inpainting posterior requires the score for a non-isotropic noising process. In both settings, the novel expressions reveal exact conditions satisfied by the posterior score, which can be used to design principled approximations.

In this section, we assume a VE noising process for simplicity of notation. For our experiments we apply both VE and VP processes; a conversion between VE and VP score functions is derived in Appendix \ref{sec:appendix_ve_vp}.

% In Proposition \ref{prop:general_score}, we present an expression for the posterior score function $\nabla \log p_t(x_t | y)$ for invertible measurement operators $A$. The score is shown to be a linear function of an auxiliary unconditional, non-isotropic noise-perturbed score function. We present results for VE due to the simpler form. For our experiments we use DDPM sampling with the VP noising process; the modified derivation is presented in the Appendix. 

\subsection{The exact posterior score for denoising}
\label{sec:exact_denoise}
When the measurement operator $A$ is the identity, the inverse problem reduces to the task of denoising, with posterior $p(x_0|y) \propto p(x_0)\,\mathcal{N}(y; x_0, \sigma^2_y I)$. It might be expected that DMs can sample from this posterior given that they are specifically trained to denoise. Indeed, several recent works \citep{wu2024principled, bruna2024posterior, xu2024provably}, show that the posterior can be sampled by a denoising process starting from time $t^\prime$ corresponding to noise level $\sigma_y$ and $x_{t^\prime} \propto y$ . While this sampling procedure has been applied to inverse problems by interleaving consistency and denoising steps \citep{wu2024principled, xu2024provably}, it does not provide direct access to the score of the noise-perturbed posterior or the noisy likelihood.

We exploit the structure of the noise-perturbed distribution to obtain a tractable expression for the posterior score (and in turn, the noisy likelihood via Eq (\ref{eqn:score_bayes})):
\begin{align*}
p_t(x_t|y) &= \int p(x_0 | y)\, \mathcal{N}(x_t; x_0, \sigma_t^2 I) \, dx_0 \\
 &\propto \int p(x_0)\, \mathcal{N}(y; x_0, \sigma_y^2 I) \, \mathcal{N}(x_t; x_0, \sigma_t^2 I) \, dx_0
\end{align*}
where we apply Bayes' rule to get the second line. By applying the rule for products of Gaussians, we arrive at a simple relationship between the perturbed posterior and prior distributions:
\begin{equation}
\label{eqn:posterior-prior}
p_t(x_t|y) \propto p_{\tilde{t}}(\tilde{x}) \mathcal{N}(y; x_t, (\sigma_y^2 + \sigma_t^2)I),
\end{equation}
where $\tilde{t}$ is defined such that $\sigma^2_{\tilde{t}} = (\sigma_y^{-2} + \sigma_t^{-2})^{-1}$ and $\tilde{x} = \tilde{x}(x_t) = \sigma_{\tilde{t}}^2 (\sigma_{y}^{-2} y + \sigma_t^{-2} x_t)$.

Finally, Proposition \ref{prop:denoising_score} gives the posterior score function for denoising, obtained by taking the score of Eq (\ref{eqn:posterior-prior}). The proof is given in Appendix~\ref{sec:proof1}.

\begin{proposition}
 For the inverse linear problem where $A = I$, the noise-perturbed posterior score function is given by
\label{prop:denoising_score}
\begin{equation}
\label{eqn:denoising_score}
\begin{split}
\nabla_{x_t} \log p_t(x_t | y) &= \sigma_{t}^{-2} \sigma_{\tilde{t}}^2 \; \nabla_{\tilde{x}} \log p_{\tilde{t}}(\tilde{x}) - (\sigma_y^2 + \sigma_t^2)^{-1}(x_t - y).
\end{split}
\end{equation}
\end{proposition}

So we can evaluate the posterior score for any $t$ as a linear function of the prior score evaluated at $\tilde{x}$ at time $\tilde{t}$. Note the following behaviors evident in Eq (\ref{eqn:denoising_score}):

\begin{itemize}
    % \item $\lim_{\sigma_y \to \infty} \nabla_{x_t} \log p_t(x_t | y) = \nabla_{x_t} \log p_t(x_t)$
    \item As $\sigma_y \to \infty$, the unconditional score $\nabla_{x_t} \log p_t(x_t)$ is recovered.
    \item For $\sigma_t \!\ll\! \sigma_y$, the RHS is approximately $\nabla_{x_t} \log p_t(x_t) - \sigma_y^{-2}(x_t - y)$, where the second term is the score of the noiseless likelihood $p(y | x_0)$, as expected.
    \item For $\sigma_t \!\gg\! \sigma_y$, $\tilde{x}(x_t) \approx y$ and $\sigma_{\tilde{t}} \approx \sigma_y$, so the arguments of the prior score function are nearly constant for the majority of time steps and, in practice, we can avoid many calls to approximate score $s_\theta(\tilde{x}, \tilde{t})$. We also note that the second term dominates for the majority of time steps; see Figure (\ref{fig:term_ratio}) Appendix~\ref{sec:tech_figures}. For sufficiently small $\sigma_y$, the linear guidance term alone, $- (\sigma_y^2 + \sigma_t^2)^{-1}(x_t - y)$, is a good approximation to the score.
    \item The noise level present in $\tilde{x}$ is $\sigma^2_{\tilde{t}}$, so the arguments of the score function are consistent with those used for the training objective of DMs. We therefore expect an off-the-shelf $s_{\theta}(\tilde{x}, \tilde{t})$ to be a reliable approximator of $\nabla_{\tilde{x}} \log p_{\tilde{t}}(\tilde{x})$.
\end{itemize}

To the best of our knowledge, Eq (\ref{eqn:denoising_score}) is a novel result that allows efficient evaluation of the posterior and noisy likelihood score functions for denoising. For methods that directly approximate the likelihood score, Eq (\ref{eqn:noisy_likelihood}), such as DPS, it can be used to evaluate the accuracy at each time step $t$ when $A=I$. This information can also be used to improve such methods. In Section \ref{sec:dps-w} we propose a simple method to determine time-dependent step sizes $\zeta_t$ to improve DPS for tasks with a large denoising character, such as colorization, random inpainting and super resolution.

\subsection{The exact posterior score for inpainting}
\label{sec:exact_inpainting}
Inpainting can be expressed as an inverse linear problem with $A=\text{diag}(d_1, \ldots, d_n)$ where $d_i \in \{0, 1\}$ determines whether pixel $i$ is masked. The posterior score for noisy inpainting, given in Proposition \ref{prop:inpainting_score}, can be derived following a procedure similar to that for denoising (see proof in Appendix \ref{sec:proof2}).

% The approach taken for denoising problems can be extended to all problems with invertible $A$. While most interesting inverse problems are ill-posed with non-invertible $A$, we can derive the score function for finite $d_i$ and take the limit as $d_i \to 0 \text{ for all masked } i$.

\begin{proposition} For inpainting problems, for which $A=\text{diag}(d_1, \ldots, d_n)$ with $d_i \in \{0, 1\}$, the noise-perturbed posterior score function is given by
\label{prop:inpainting_score}
\begin{equation}
\label{eqn:inpainting_score}
\begin{split}
\nabla_{x_t} \log p_t(x_t | y) &= \sigma_{t}^{-2} \Sigma_{\tilde{t}} \; \nabla_{\tilde{x}} \log p_{\Sigma_{\tilde{t}}}(\tilde{x}) - (\sigma_y^2 + \sigma_t^2)^{-1}A(x_t - y),
\end{split}
\end{equation}
where $\Sigma_{\tilde{t}} = (\sigma^{-2}_{y}A + \sigma^{-2}_{t}I)^{-1}$, $\tilde{x} = \Sigma_{\tilde{t}} (\sigma_{y}^{-2} Ay + \sigma_t^{-2} x_t)$ and the non-isotropic score function
\begin{equation}
\label{eq:non-isotropic-score}
\nabla_{\tilde{x}} \log p_{\Sigma_{\tilde{t}}}(\tilde{x}) = \int p(x_0)\, \mathcal{N}(x_t; x^\prime, \Sigma_{\tilde{t}}) \, dx_0.
\end{equation}
\end{proposition}

Given the score function for a noising process that has a different noise level for each pixel, we can compute the exact posterior score using Eq (\ref{eqn:inpainting_score}). Such non-isotropic score functions are not commonly available, but there is recent work in training such models and performing conditional sampling with non-isotropic denoising processes \citep{gerdes2024gud}. Eq (\ref{eqn:inpainting_score}) shows how to sample from the exact posterior using a non-isotropic score in an isotropic denoising process.

In the Experiments section, we demonstrate the validity of the posterior score for inpainting on a toy problem for which the non-isoptropic score function is analytically tractable.

Analysis of Eq (\ref{eqn:inpainting_score}) reveals the following properties of the exact posterior score:
\begin{itemize}
    \item For $\sigma_y \to 0$, $\tilde{x}$ and $\Sigma_{\tilde{t}}$ define a noising process that only noises the masked pixels. This is an intuitive result; we would expect a model trained on this process to solve noiseless inpainting tasks.
    \item Despite the intractability of the score overall, the components corresponding to unmasked pixels are given exactly by $- \sigma_t^{-2}A(x_t - y)$ for noiseless inpainting.
    \item For $\sigma_y > 0$, Eq (\ref{eqn:inpainting_score}) describes the exact balancing between noise levels in the non-isotropic score for masked and unmasked pixels at each time step required to sample from the posterior.
    \item For $\sigma_t \gg \sigma_y$, $A \nabla_{x_t} \log p_t(x_t | y) \approx - (\sigma_y^2 + \sigma_t^2)^{-1}A(x_t - y)$. As noted in \ref{sec:exact_denoise}, and demonstrated in Figure~\ref{fig:term_ratio} in Appendix~\ref{sec:tech_figures} for denoising, the linear guidance term dominates for large $\sigma_t$.
    \item For $\sigma_t \ll \sigma_y$, $\nabla_{\tilde{x}} \log p_{\Sigma_{\tilde{t}}}(\tilde{x}) \approx \nabla_{x_t} \log p_t(x_t)$ and the noisy likelihood score can be approximated by the simple guidance term $- \sigma_y^{-2}A(x_t - y)$.
    \item For $\sigma_y \approx \sigma_t$, for the typically low values of $\sigma_y$, the anisotropy of $\Sigma_{\tilde{t}}$ is small in absolute terms, and we expect the isotropic score function in the denoising posterior score expression to provide a good approximation of the non-isotropic score for unmasked pixel dimensions.
\end{itemize}

As a result, we expect the denoising trajectories of unmasked pixels by the posterior score for denoising, Eq (\ref{eqn:denoising_score}), and inpainting, Eq (\ref{eqn:inpainting_score}), to be identical for $\sigma_y=0$ and approximately equal for $\sigma_y > 0$.

\subsection{DPS-w: improving DPS with time-dependent step size}
\label{sec:dps-w}
In practice, the DPS method replaces the analytically-derived step size of $\sigma_y^{-2}$ with the factor $\zeta_t = \zeta^\prime/\|y - A \hat{x}_0(x_t)\|$. The division by the norm is motivated \citep{DPS, LGD} by the expected increase in error for larger $\sigma_t$; the norm will generally be larger earlier on in the denoising process. The constant $\zeta^\prime$ is a task-specific hyperparameter that is determined empirically. Improvements to DPS are generally motivated by error analysis of toy problems (e.g. \citep{LGD, thaker2024frequency}) or theoretical grounds such as manifold preservation \citep{mcg, he2023manifold, DSG}. For real-world problems, reliable ground truth data is generally unavailable for validation of approximate score functions. With access to the tractable posterior score, Eq (\ref{eqn:denoising_score}), we can evaluate approximate scores for the case of denoising, and even improve them on-the-fly.

We can combine Eqs (\ref{eqn:denoising_score}) and (\ref{eqn:score_bayes}) to yield an expression for the noisy likelihood score,
\begin{align}
    \nabla_{x_t}\log p_t(y | x_t) &= \nabla_{x_t}\log p_t(x_t | y) - \nabla_{x_t}\log p_t(x_t) \nonumber \\
    \label{eqn:exact_noisy_likelihood}&=\sigma_{t}^{-2} \sigma_{\tilde{t}}^2 \; \nabla_{\tilde{x}} \log p_{\tilde{t}}(\tilde{x}) - \nabla_{x_t}\log p_t(x_t) - (\sigma_y^2 + \sigma_t^2)^{-1}(x_t - y).
\end{align}
Substituting the trained score $s_\theta$ for both prior score invocations on the RHS of Eq (\ref{eqn:exact_noisy_likelihood}), we define
\begin{equation}
\label{eqn:ref_noisy_likelihood}
s_\theta(y | x_t) = \sigma_{t}^{-2} \sigma_{\tilde{t}}^2 \; s_\theta(\tilde{x}, \tilde{t}) - s_\theta(x_t, t) - (\sigma_y^2 + \sigma_t^2)^{-1}(x_t - y),
\end{equation}
which can be compared directly to the DPS approximation, $s_\text{DPS}(y|x_t, A=I)$, where
\begin{equation}
\label{eqn:s_dps}
s_\text{DPS}(y | x_t, A) = - \zeta_t \nabla_{x_t} \| y - A\hat{x}_0(x_t) \|^2.
\end{equation}
We propose the DPS-w method, which replaces the hyperparameter $\zeta_t$ at each time step with the weight $w_t$ that minimizes the MSE of the DPS score for the reference task of denoising ($A=I$):
\begin{equation}
w_t = \frac{s_{\theta}(y | x_t)\cdot s_\text{DPS}(y|x_t, A=I)}{\|s_\text{DPS}(y|x_t, A=I)\|^2},
\end{equation}
where $\cdot$ is the dot product. While the weight $w_t$ is optimized for the pure denoising case, it can be applied to DPS for general inverse problems. For problems with a high-degree of denoising character, such as colorization, random inpainting and super-resolution, we expect the $w_t$ to be informative and improve the DPS trajectory. For a given inverse problem, a reference denoising task is chosen and used to compute $w_t$. The reference task for inpainting is denoising of the unmasked pixels, for colorization it is denoising of the noised grayscale image and for super resolution it is denoising of the adjoint-upsampled measurement. Full details and algorithm are given in Appendix \ref{sec:dps-w-sr}. In the experimental section we show that, despite its simplicity, DPS-w provides a significant improvement for these tasks, competitive with more sophisticated state-of-the-art methods at varying levels of measurement noise.

The values of ${w_t}$ for a typical denoising task are presented in Figure~\ref{fig:wt_plot} in Appendix \ref{sec:tech_figures}. It is clear that the reciprocal error norm heuristic employed by DPS leads to step sizes that far too large for $A=I$. The PGDM \citep{song2023pseudoinverse} method also introduced time-dependent time steps based on a heuristic motivated by pure denoising guidance, approaching $1$ for large $\sigma_t$ and $0$ for small $\sigma_t$. As discussed in \ref{sec:image_exps}, large, early step sizes are a feature of DSG guidance. In contrast, the $w_t$ of DPS-w, show the opposite behavior, starting out very small and becoming much larger towards the end of the trajectory.

\section{Experiments}
In the \ref{sec:toy_problem}, we compare the performance of the exact posterior score function introduced in the Methods section, with DPS and DPS-w on a toy problem. In \ref{sec:image_exps}, we evaluate the DPS-w method on two standard benchmarking image datasets for inverse linear problems: denoising, colorization, random inpainting and super resolution.

\subsection{Toy problem: two-dimensional double well}
\label{sec:toy_problem}

We introduce a double-well model system represented by a bivariate Gaussian mixture distribution, $p_\text{DW}(x^{(0)}, x^{(1)})$. The two Gaussians are equally weighted and have means $\boldsymbol{\mu}_1 = (-2.0, 2.4)$, $\boldsymbol{\mu}_2 = (1.5, 0.0)$ and standard deviations $\boldsymbol{\sigma}_1 = (0.5, 0.6)$, $\boldsymbol{\sigma}_2 = (0.3, 0.45)$, respectively. The log-probability of the distribution is visualized in Figure~\ref{fig:DW}, Appendix \ref{sec:tech_figures}, along with one hundred random samples. A Gaussian mixture is a convenient choice since it has an analytic noise-perturbed score function for general Gaussian noise, including the non-isotropic noise in Eq (\ref{eq:non-isotropic-score}). 

Consider the problem of computing the free energy profile along the first dimension (horizontal axis of Figure~\ref{fig:DW}(a), which is the negative logarithm of the marginal
\begin{equation}
p(x^{(0)}) = \int p_\text{DW}(x^{(0)}, x^{(1)}) dx^{(1)}.
\end{equation}
While the integral is tractable in this case, we aim to solve the problem by sampling with a denoising process. Simply sampling from the unconditional score will yield good representation of the minima, but very little coverage of the low-probability regions, such as the transition barrier between the two wells. A popular technique from the field of molecular dynamics, Umbrella Sampling \citep{umbrella}, addresses this issue through use of harmonic bias potentials centered at various positions along $x^{(0)}$. The biased samples are processed with the weighted-histogram analysis technique (WHAM) \citep{WHAM} to remove the biases in the computed integral.

For this problem, Umbrella Sampling turns out to be identical to noisy inpainting, with mask $A=\text{diag}(1, 0)$ and noise level $\sigma_y$ governing the width of a bias window. We perform sampling at multiple positions using the exact posterior score for inpainting given by Eq (\ref{eqn:inpainting_score}) (see Figure~\ref{fig:DW}(b)), and approximations DPS and DPS-w, and compute the corresponding free energy profiles. Figure~\ref{fig:DW}(c) shows that the exact posterior score correctly represents the harmonic bias potential, leading to a highly accurate free energy profile when unbiasing with WHAM. DPS is too flexible in the $x^{(1)}$ dimension, and too restrictive in $x^{(0)}$, requiring twice as many bias windows and producing a flat profile that lacks any characteristic of the ground truth. On the other hand, DPS-w captures the correct shape, albeit with an underestimation of the depth of the first well.

\subsection{Inverse linear problems on images}
\label{sec:image_exps}
We conduct quantitative and qualitative evaluation of the DPS-w method against benchmark methods DPS, DSG and DAPS, on the FFHQ \citep{karras2019style} and ImageNet \citep{deng2009imagenet} datasets with resolution $256\times256$. We use the first 200 images from the FFHQ 1k validation set and a random sample of 200 images from the ImageNet 1k validation set. Following DPS \citep{DPS}, the same pre-trained, unconditional DDPM models are used for all tasks and benchmark methods: for FFHQ we use the model released with DPS; for ImageNet we use the model from \citep{dhariwal2021diffusion}. For evaluation metrics, we compute LPIPS \citep{zhang2018unreasonable} with the replace-pooling option enabled, SSIM and PSNR. Following the implementation in \citep{DAPS}, the metrics are computed with the piq \citep{piq} library with data normalized to the range $[0, 1]$. All methods are run with $1000$ steps, except the $100$-step version of DPS-w.

Measurement noise $\sigma_y$ is added according to Eq (\ref{eqn:forward_model}), i.e. the measurement operator is applied to the normalized image first, followed by noise. We run experiments for $\sigma_y=0.01$, $0.05$ and $0.1$. The benchmarking experiments in \citep{DPS}, \citep{DAPS} and \citep{DSG} are carried out for $\sigma_y=0.05$. It is important to highlight an ambiguity in recent publications, particularly works that extend the code base of DPS (including DSG and our work). The DPS paper reports that images are normalized to the range $[0, 1]$, but in the code the range is actually $[-1, 1]$. The true noise level in the $[0, 1]$ pixel space is therefore $\sigma_y/2$ and it is not clear whether the value of $\sigma_y$ used for experiments was adjusted accordingly. We assume that the noise level corresponds to the data normalized to $[-1, 1]$; our benchmarking experiments are consistent with this for all methods. We also include $\sigma_y=0.1$ for some experiments to cover both scenarios.

For each method, we use the same published hyperparameters where available for a given task, and borrow hyperparamers from a related task if not (e.g. inpainting with 70\% vs 92\% masking). Since all benchmark methods were developed for $\sigma_y=0.05$, the purpose of testing on different noise levels is to evaluate sensitivity and broad applicability of the parameters, rather than assessing a method's full potential. When comparing to DPS-w, we highlight the tasks to which each method was fine tuned and should therefore be the most competitive. DPS-w has no hyperparameters, except for the single $w_\text{max}$ parameter in the case of super resolution, which is fit to a single image per dataset and not fine tuned for different noise levels (see Appendix for more details).

\textbf{Random inpainting.} We run experiments for random inpainting with 40\%, 70\%, and 92\% of pixels removed. With decreasing masking probability, the inpainting problem resembles pure denoising more closely.

\textbf{Super resolution.} 
The image resolution is reduced to $64\times64$ with bicubic downsampling.

Table \ref{main_table} in Appendix \ref{sec:appendix_tables} presents the quantitative benchmarking results for inpainting and super resolution tasks, with $\sigma_y = 0.05$ on both FFHQ and ImageNet. Overall, DPS-w is competitive on these tasks, out performing DAPS on most tasks and metrics, and competitive with DSG; while DSG has the best LPIPS evaluation across all tasks, DPS-w has stronger results for the SSIM and PSNR metrics. DPS-w significantly improves on DPS in all cases. Interestingly, the $100$-step version of DPS-w appears to trade off a small loss in LPIPS performance for gain in SSIM and PSNR; beating DPS-w with $1000$ steps in most cases. Confidence intervals for these results are provided in Tables \ref{main_table_ci_ffhq} and \ref{main_table_ci_imagenet} in Appendix \ref{sec:appendix_tables}.

By scaling the DPS-w step size by a factor of $\sqrt{d_m}$, where $d_m$ is the number of masked pixels, random inpainting performance on ImageNet was improved to match DSG. See Appendix \ref{sec:enhanced_guidance} for results and discussion.

For the ImageNet super resolution task, all methods had a mix of success and failure cases, with failures including blurry or distorted images. While DPS-w performed well quantitatively on the super resolution task, we noticed a higher number of blurry samples compared to DSG. More thorough parametrization of the DPS-w reference denoising task for super resolution, beyond $w_\text{max}$, would likely improve these results, but is against the spirit and scope of this work.

\textbf{Other noise levels.}
In DPS, the noise level is absorbed into $\zeta^\prime$ and there is no other reference to $\sigma_y$. Similarly, the DAPS method was developed for a wide range of tasks exclusively at $\sigma_y=0.05$, and the noise level itself is replaced with a tuned hyperparameter. On the other hand, DPS-w has a concrete dependency on $\sigma_y$ via the denoising score function, and DSG has adaptable step sizes and robust parameters that are claimed to be broadly applicable across tasks. Additional benchmarking results for $\sigma_y = 0.01$ and $0.1$ can be found in Tables \ref{table:sigma_0.01} and \ref{table:sigma_0.1} in Appendix \ref{sec:appendix_tables}, respectively. DPS-w outperforms the benchmark methods across almost all tasks for both $\sigma_y = 0.01$ and $0.1$, demonstrating the value of the denoising posterior score reference as a way to avoid over-dependence on hyperparameters. For applications where the noise level is not known, $\sigma_y$ can be treated as a problem-specific hyperparameter.

\textbf{Denoising.} The results for the denoising task with $\sigma_y = 0.05$ using the exact posterior score, Eq~(\ref{eqn:denoising_score}), DPS, DPS-w, and DPS-w with 100 steps are given in Table \ref{table:denoising}. For DPS, we use the random inpainting configuration and set masking probability to zero. As expected, the exact score gives superior performance to the DPS reference, which has not been specifically optimized for denoising. DPS-w with its optimized step sizes closely matches the exact results. DPS-w with 100 steps is only slightly worse in the LPIPS metric. In Appendix \ref{sec:nec_conds_posterior}, we assess how well the exact denoising posterior satisfies necessary conditions of a true posterior sampler when using the unconditional score of the FFHQ model.

\begin{table}[ht]
\caption{Quantitative evaluation of image denoising on FFHQ.}
\label{table:denoising}
\centering
% \begin{small}
    \begin{tabular}{lccc}
        \toprule
        Method & LPIPS ($\downarrow$) & SSIM ($\uparrow$) & PSNR ($\uparrow$) \\
        \midrule
        Exact & 0.033 & 0.954 & 37.172 \\
        DPS & 0.117 & 0.898 & 32.344 \\
        DPS-w & 0.048 & 0.954 & 37.181 \\
        \bottomrule
    \end{tabular}
% \end{small}
\end{table}

We also examine denoising trajectories for a FFHQ image (\autoref{fig:appendix:denoise_trajectory} in Appendix \ref{sec:further-image}) for exact, DPS, DPS-w and DSG methods. Due to the small weights $w_t$ for large $t$, DPS-w is seen to develop high-level features slightly later than DPS. Conversely, as highlighted by the authors \citep{DSG}, DSG develops high-level features very early in the denoising process, and it is suggested that this ``more effective'' guidance enables rapid sampling. By comparing to the exact trajectory, which is slower to develop features, we can see that the impressive performance of DSG guidance is not due to an improved approximation of the score function. Additionally, the ability of DPS-w to sample with fewer steps without hyperparameter tuning, shows that strong, early guidance is not essential for rapid sampling.

\begin{figure}[ht]
\centering
\includegraphics[width=0.8\columnwidth]{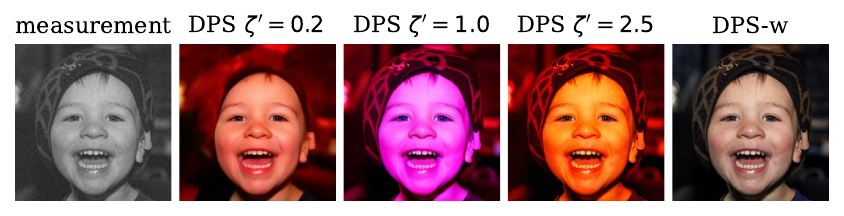}
\caption{DPS samples for various $\zeta^\prime$ values on the image colorization task with $\sigma_y=0.1$. DPS is unable to produce realistic results across a range of hyperparameter values, without loss of structural features ($\zeta^\prime=0.2$). DPS-w yields high-quality samples despite having no tunable parameters.}
\label{fig:color_scan}
\end{figure}

\textbf{Colorization.} The measurement operator takes a weighted average of the color channels per pixel to obtain a grayscale image. The values are repeated into three channels so that the shape of the image tensor is unchanged. Figure~\ref{fig:color_scan} shows samples from DPS from a hyperparameter scan for an FFHQ image, compared to DPS-w. DPS is unable to generate high quality samples across a range of hyperparameter values. For large $\zeta^\prime > 0.2$, DPS generates strongly color-tinted images; at lower guidance step size, the unconditional score helps obtain a more realistic color distribution, but at the cost of losing structural features in the image. The time-dependent $w_t$ of DPS-w enable a more flexible trade off, avoiding a commitment to color features too early in the trajectory, while giving the prior more weight towards the end. More samples are presented for qualitative evaluation on FFHQ and ImageNet in Figures~\ref{fig:ffhq_color} and \ref{fig:imagenet_color} in Appendix \ref{sec:further-image}, respectively. ImageNet proves to be a more challenging dataset for this task: while most of the ten presented samples are reasonable, there are a couple of failure cases with color bleeding, or patching.
% Colorization samples for ImageNet are presented in Fig (\ref{imagenet_color}). For $\sigma_y = 0.1$, some samples have local patches of consistent coloring with drastic changes between boundaries. This could be due to the inaccuracies from unconditional score for the grayscale reference images used for DPS-w. A higher $\sigma_y$ will lead to earlier mixing of the grayscale measurement with the noisy image $x_t$, as well as give more overall weighting to the prior score. The patching issue seems to be resolved by setting $\simga_y=0.2$

\section{Limitations}
While the value of the exact posterior score is demonstrated through the success of the DPS-w method for colorization, inpainting and super resolution tasks, the benchmark methods are successfully applied to a much broader set of inverse problems. Incorporation of the tractable score into a more generalized framework for training-free guidance is the natural next step. Also, it was not explored whether the approach can be extended to guidance of latent diffusion models\citep{rombach2022high}.

\section{Conclusions}
We present a tractable expression for the denoising posterior score function, in terms of the unconditional score function of Diffusion Models, and demonstrate its value in the analysis and improvement of training-free guidance methods. To further demonstrate the utility of the exact posterior score, we propose DPS-w, a hyperparameter-free method that computes locally-optimal step sizes for DPS denoising tasks and transfers them to other inverse problems: colorization, inpainting and super resolution. The method is shown to be competitive with two of the latest start-of-the-art methods, DAPS and DSG, with little computational overhead or, in the case of the $100$ time step version, much improved sampling times. We hope that this work inspires new approaches to the development of robust training-free guidance methods.

% \bibliographystyle{unsrtnat}
% \bibliography{references}

\bibliographystyle{unsrtnat}
\bibliography{references}

\clearpage
\begin{center}
    {\Large \textbf{Exploiting the Exact Denoising Posterior Score in Training-Free Guidance of Diffusion Models}}\\[3ex]
{\Large \textbf{Appendix}}
\appendix

\end{center}

\section{Proofs}

\subsection{Proof of Proposition~\ref{prop:denoising_score}}
\label{sec:proof1}
\begin{proposition} For the inverse linear problem where $A = I$, the noise-perturbed posterior score function is given by
\label{prop:denoising_score_appendix}
\begin{equation}
\label{eqn:denoising_score_appendix}
\begin{split}
\nabla_{x_t} \log p_t(x_t | y) &= \sigma_{t}^{-2} \sigma_{\tilde{t}}^2 \; \nabla_{\tilde{x}} \log p_{\tilde{t}}(\tilde{x}) \\
&\quad - (\sigma_y^2 + \sigma_t^2)^{-1}(x_t - y).
\end{split}
\end{equation}
\end{proposition}

\textbf{Proof:} We start by considering the broader subset of inverse problems for which the measurement operator $A$ is invertible and the likelihood is

\begin{equation}
    \label{eqn:likelihood_appendix}
    p(y|x_0) = \mathcal{N}(y; Ax_0, \sigma^2_y I).
\end{equation}

For a VE noising process with total noise level $\sigma_t$ at time step $t$, the noise-perturbed posterior distribution is

\begin{equation}
p_t(x_t|y) = \int p(x_0 | y)\, \mathcal{N}(x_t; x_0, \sigma_t^2 I) \, dx_0,
 % &\propto \int p(x_0)\, \mathcal{N}(y; x_0, \sigma_y^2 I) \, \mathcal{N}(x_t; x_0, \sigma_t^2 I) \, dx_0
\end{equation}

substituting in Eq (\ref{eqn:likelihood_appendix}) and the result from Bayes' rule $p(x_0 | y) \propto p(x_0) p(y | x)$ gives

\begin{equation}
\label{eqn:two_gaussians}
p_t(x_t|y) \propto \int p(x_0)\, \mathcal{N}(y; A x_0, \sigma_y^2 I) \, \mathcal{N}(x_t; x_0, \sigma_t^2 I) \, dx_0.
\end{equation}

We will manipulate the form of the Normal density functions, losing the probabilistic interpretation in terms of random variables, so we introduce $n$ to represent the normalized Gaussian density function to avoid confusion. Due to the symmetry of the Gaussian function, $n(y ; Ax_0, \sigma^2_y I) = n(Ax_0; y, \sigma^2_y I)$. Completing the square in the exponent gives the change of variables:

\begin{equation}
\label{eqn:rearrange}
    n(y; Ax_0, \sigma^2_y I) \propto n(x_0; A^{-1}y, (\sigma^{-2}_yA^TA)^{-1}).
\end{equation}

Substituting Eq (\ref{eqn:rearrange}) into Eq (\ref{eqn:two_gaussians}),

\begin{equation}
    p_t(x_t|y) \propto \int p(x_0)\, n(x_0; \mu_A, \Sigma_A)\, n(x_t; x_0, \sigma_t^2 I) \, dx_0,
\end{equation}

with $\mu_A = A^{-1}y$ and $\Sigma_A = (\sigma^{-2}_yA^TA)^{-1}$.

Note that $x_t$ is a constant in the integration with respect to $x_0$ so, making use of symmetry again, we treat $x_t$ as the mean, $n(x_t; x_0, \sigma_t^2 I) = n(x_0; x_t, \sigma_t^2 I)$. The integrand can then be simplified through use of the rule for Gaussians products,

\begin{equation}
    n(x_0; \mu_A, \Sigma_A)\, n(x_0; x_t, \sigma_t^2 I) = n(x_t; \mu_A, \Sigma_A + \sigma^2_t I)\,n(x_0, \mu, \Sigma),
\end{equation}

where
\begin{equation}
    \Sigma = (\Sigma^{-1}_A+\sigma_t^{-2} I)^{-1}
\end{equation}
and
\begin{align}
\mu = \mu(x_t) &= \Sigma(\Sigma^{-1}_A A^{-1} y + \sigma_t^{-2} x_t)\nonumber \\
&= \Sigma(\sigma^{-2}_yA^T y + \sigma_t^{-2} x_t),
\end{align}

to give

\begin{equation}
\label{eqn:new_form_posterior}
    p_t(x_t|y) \propto n(x_t; \mu_A, \Sigma_A + \sigma^2_t I) \, \int p(x_0)\,
      n(\mu; x_0, \Sigma) \, dx_0.
\end{equation}

The integral is once again a convolution of with a Gaussian density, but now of the prior $p(x_0)$ with a non-isotropic Gaussian, $n(\mu; x_0, \Sigma)$. We define the non-isotropic noise-perturbed distribution $p_\Sigma(x) \triangleq \int p(x_0) n(x; x_0, \Sigma) d x_0$.

Introducing the constant of proportionality $c$ and taking the logarithm of both sides:

\begin{equation}
    \log p_t(x_t|y) = \log n(x_t; \mu_A, \Sigma_A + \sigma^2_t I) + \log p_\Sigma(\mu(x_t)) + \log c.
\end{equation}

Finally, taking the gradient gives the score

\begin{align}
\label{eqn:full_form_score}
    \nabla_{x_t} \log p_t(x_t|y) &= \nabla_{x_t}\log n(x_t; \mu_A, \Sigma_A + \sigma^2_t I) + \nabla_{x_t}\log p_\Sigma(\mu(x_t)) \nonumber \\
    &=  \sigma^{-2}_t (\sigma^{-2}_yA^TA+\sigma^{-2}_tI)^{-1} \, \nabla_{x_t} \log_\mu p_\Sigma(\mu) \nonumber \\
    &\hspace{1em}- (\sigma^2_y (A^TA)^{-1} + \sigma^2_t I)^{-1}(x_t - A^{-1}y)
\end{align}

and the denoising score function, Eq (\ref{eqn:denoising_score_appendix}), is obtained by setting $A=I$.

\subsection{Proof of Proposition \ref{prop:inpainting_score}}
\label{sec:proof2}

\begin{proposition} For inpainting problems, for which $A=\text{diag}(d_1, \ldots, d_n)$ with $d_i \in \{0, 1\}$, the noise-perturbed posterior score function is given by
\label{prop:inpainting_score_appendix}
\begin{equation}
\label{eqn:inpainting_score_appendix}
\begin{split}
\nabla_{x_t} \log p_t(x_t | y) &= \sigma_{t}^{-2} \Sigma_{\tilde{t}} \; \nabla_{\tilde{x}} \log p_{\Sigma_{\tilde{t}}}(\tilde{x}) \\
&\quad - (\sigma_y^2 + \sigma_t^2)^{-1}A(x_t - y),
\end{split}
\end{equation}
where $\Sigma_{\tilde{t}} = (\sigma^{-2}_{y}A + \sigma^{-2}_{t}I)^{-1}$, $\tilde{x} = \Sigma_{\tilde{t}} (\sigma_{y}^{-2} Ay + \sigma_t^{-2} x_t)$ and the non-isotropic score function

\begin{equation}
\label{eq:non-isotropic-score-appendix}
\nabla_{\tilde{x}} \log p_{\Sigma_{\tilde{t}}}(\tilde{x}) = \int p(x_0)\, \mathcal{N}(x_t; x^\prime, \Sigma_{\tilde{t}}) \, dx_0.
\end{equation}

\end{proposition}

Consider a specific form of the measurement operator where $A = \text{diag}(e_1, \ldots, e_n)$ with $e_i \in \{1, \delta\}$ and $\delta > 0$. This A is an invertible, positive definite matrix, and therefore Eq (\ref{eqn:full_form_score}) applies. In the limit as $\delta \to 0$, $A$ is the inpainting measurement operator with $e_i \in \{0, 1\}$. The noise-perturbed posterior score for inpainting is therefore obtained by taking the limit of Eq (\ref{eqn:full_form_score}) as $\delta \to 0$, giving Eq (\ref{eqn:inpainting_score_appendix}) in the Proposition.

\section{Implementation details}

\subsection{Converting between VE and VP processes}
\label{sec:appendix_ve_vp}

The difference in the forward-noising posterior distribution $p(x_t|x_0)$ for VE versus VP processes leads to differences in the noise-perturbed distribution $p_t(x_t)$. Recall the posterior distributions for VE and VP, respectively:

\begin{align}
    q_t^{\text{VE}}(x_t|x_0) &= \mathcal{N}(x_t; x_0, \sigma^2_t I),\\
    q_t^{\text{VP}}(x_t|x_0) &= \mathcal{N}(x_t; \sqrt{\bar{\alpha}_t}x_0, (1 - \bar{\alpha}_t) I),
\end{align}

The score functions in Proposition \ref{prop:denoising_score} and Proposition \ref{prop:inpainting_score} are derived for VE, for which $p_t(x_t)$ is the convolution of the conditional data distribution with $q_t^{\text{VE}}$. With a change of variables, the VP posterior density can be written as a VE posterior, evaluated with a modified $x_t$ and variance:

\begin{equation}
    n(x_t; \sqrt{\bar{\alpha}_t}x_0, (1 - \bar{\alpha}_t) I) \propto  n(\bar{\alpha}_t^{-1/2}x_t; x_0, \frac{1-\bar{\alpha}_t}{\bar{\alpha}_t} I).
\end{equation}

Therefore, by setting $\sigma^2_t \coloneqq (1 - \bar{\alpha}_t)/\bar{\alpha}_t$ and $x_t \coloneqq \bar{\alpha}_t^{-1/2}x_t$, $q_t^{\text{VE}}$ is equivalent to $q_t^{\text{VP}}$, ignoring a proportionality constant. Applying this substitution to Eq~(\ref{eqn:new_form_posterior}) and setting $A=I$ gives

\begin{equation}
\label{eqn:new_form_posterior_VP}
    p^\text{VP}_t(x_t|y) \propto n(\bar{\alpha}_t^{-1/2}x_t; y, (\sigma_y^2 + \frac{1-\bar{\alpha}_t}{\bar{\alpha}_t})I) \, \int p(x_0)\,
      n(\tilde{x}; x_0, \sigma_{\tilde{t}}^2 I) \, dx_0,
\end{equation}

where
\begin{equation}
    \sigma_{\tilde{t}}^2 = (\sigma_y^{-2}+\frac{\bar{\alpha}_t}{1-\bar{\alpha}_t})^{-1}
\end{equation}
and
\begin{equation}
\tilde{x} = \tilde{x}(x_t) = \sigma_{\tilde{t}}^2(\sigma_y^{-2}  y + \frac{\bar{\alpha}^{1/2}_t}{1-\bar{\alpha}_t} x_t).
\end{equation}

To make the convolution integral in Eq~(\ref{eqn:new_form_posterior_VP}) resemble a noise-perturbed distribution for VP process, apply another change of variables, $n(\tilde{x}; x_0, \sigma_{\tilde{t}}^2) \propto n(\sqrt{\bar{\alpha}_t}\tilde{x}; \sqrt{\bar{\alpha}_t}x_0, \bar{\alpha}_t\sigma_{\tilde{t}}^2)$:

\begin{align}
\label{eqn:new_form_posterior_VP_resolved}
    p^\text{VP}_t(x_t|y) &\propto n(\bar{\alpha}_t^{-1/2}x_t; y, (\sigma_y^2 + \frac{1-\bar{\alpha}_t}{\bar{\alpha}_t})I) \, \int p(x_0)\,
      n(\sqrt{\bar{\alpha}_t}\tilde{x}; \sqrt{\bar{\alpha}_t}x_0, \bar{\alpha}_t\sigma_{\tilde{t}}^2) \, dx_0 \nonumber \\
      &\propto  n(\bar{\alpha}_t^{-1/2}x_t; y, (\sigma_y^2 + \frac{1-\bar{\alpha}_t}{\bar{\alpha}_t})I) \, p_{\tau}^\text{VP}(\sqrt{\bar{\alpha}_t}\tilde{x}),
\end{align}
where $\tau$ is defined such that $1-\bar{\alpha}_\tau = \bar{\alpha}_t\sigma_{\tilde{t}}^2$. The score is therefore given by

\begin{align}
\label{eqn:new_form_posterior_VP_resolved_2}
    \nabla_{x_t} \log p_t^{\text{VP}}(x_t|y) & = \nabla_{x_t} \log p_{\tau}^{\text{VP}}(\sqrt{\bar{\alpha}_t}\tilde{x}) + \nabla_{x_t} \log n(\bar{\alpha}_t^{-1/2}x_t; y, (\sigma_y^2 + \frac{1-\bar{\alpha}_t}{\bar{\alpha}_t})I) \nonumber \\
    & = \frac{\bar{\alpha}_t}{1-\bar{\alpha}_t}\sigma^2_{\tilde{t}} \nabla_{\mu} \log p^{\text{VP}}_{\tau}(\mu) - \frac{1}{\bar{\alpha}_t}(\sigma_y^2 + \frac{1-\bar{\alpha}_t}{\bar{\alpha}_t})^{-1}(x_t - \sqrt{\bar{\alpha}_t}y)
    ,
\end{align}
where \(\mu = \sqrt{\bar{\alpha}_t}\tilde{x}\).

Finally, with a DDPM approximate score function, $s_\theta$, we can approximate the posterior score function for denoising as:

\begin{equation}
\label{eqn:new_form_posterior_VP_resolved_3}
    \nabla_{x_t} \log p_t^{\text{VP}}(x_t|y)  \approx \frac{\bar{\alpha}_t}{1-\bar{\alpha}_t}\sigma^2_{\tilde{t}} s_\theta(\sqrt{\bar{\alpha}_t} \tilde{x}, \tau) - \frac{1}{\bar{\alpha}_t}(\sigma_y^2 + \frac{1-\bar{\alpha}_t}{\bar{\alpha}_t})^{-1}(x_t - \sqrt{\bar{\alpha}_t}y).
\end{equation}

\subsection{Computational cost of DPS-w}
In a naive implementation of DPS-w, the score model is called an additional time every time step, to compute the reference denoising score. As suggested in Section \ref{sec:exact_denoise}, this is unnecessary because $\tilde{x}$ is the harmonic mean of the measurement $y$ and $x_t$, with weights $\sigma^2_y$ and $\sigma^2_t$. For $\sigma_y = 0.05$, the relative weight of $y$ at time step $100$ is approximately $98\%$. Therefore, a reasonable approximation is to compute and cache the score for $\tilde{x} = y$ and $\sigma_{\tilde{t}} = \sigma_y$ at the start of a sampling procedure, and reuse for the first $900$ time steps when sampling with $1000$ steps. As shown in Figure~\ref{fig:term_ratio} in Appendix~\ref{sec:tech_figures}, for $\sigma_y=0.05$ and $t=100$, the term requiring computation of $s_\theta(\tilde{x}, \sigma_{\tilde{t}})$ contributes less than $10\%$ to the total posterior score, further improving the suitability of this caching strategy. The DPS-w method therefore requires at most $10\%$ more forward evaluations of the neural network score model than DPS when using $1000$ time steps for both methods. For the sake of simplicity, our benchmark experiments were run using the naive implementation.

The DPS norm gradient requires backpropagation through the score model. Using DPS-w for colorization and super resolution require evaluation of an extra DPS gradient, but both DPS gradients can be computed with a single backpropagation step with negligible additional cost.

Number of forward evaluations (NFEs) of the neural network is a common unit for quantifying the computational overhead of posterior sampling with diffusion models. DPS-w does not introduce other bottlenecks, so NFEs is a suitable measure of cost. DPS-w has a cost of ~1100 NFEs to perform sampling with 1000 diffusion steps. DPS, DAPS and DSG benchmark methods were also run with 1000 steps and therefore 1000 NFEs per sample.

\subsection{Definition of problem-specific reference tasks for DPS-w}
\label{sec:dps-w-sr}
The procedure for posterior sampling with DPS-w is given in Algorithm~\ref{alg:dps-w}. It is a reproduction of the DPS\citep{DPS} algorithm for Gaussian noise, with the DPS step size $\zeta_t$ replaced by $w_t$ calculated for a reference denoising task. The number of time steps $N$ is set to $1000$ in all cases, expect when DPS-w with run with $100$ time steps. The DDPM\citep{ddpm} schedule parameters $\bar{\alpha}_t$, $\alpha_t$, and learned variances $\tilde{\sigma}_t^2$ are as defined in the DPS paper.

Task-specific definitions of $w_t$ for denoising, super resolution, inpainting, and colorization are given in subsections \ref{sec:wt_denoising}, \ref{sec:wt_sr}, \ref{sec:wt_inpaint}, and \ref{sec:wt_color}, respectively.

\begin{algorithm}[H]
\label{alg:dps-w}
\caption{DPS-w}
\KwRequire{$N$, $y$, $A$, $\{\tilde{\sigma}_t\}_{t=1}^N$}
$\mathbf{x}_N \sim \mathcal{N}(0, I)$

\For{$t = N-1$ \KwTo $0$}{
    $\hat{x}_0 \leftarrow \frac{1}{\sqrt{\bar{\alpha}_t}} \left( x_t + (1 - \bar{\alpha}_t) s_\theta(x_t, t) \right)$
    
    $z \sim \mathcal{N}(0, I)$

    $x_{i-1}' \leftarrow \sqrt{\alpha_t} \frac{(1 - \bar{\alpha}_{i-1})}{1 - \bar{\alpha}_t} x_t + 
    \frac{\sqrt{\bar{\alpha}_{i-1}} \beta_t}{1 - \bar{\alpha}_t} \hat{x}_0 + \tilde{\sigma}_t z$

    \texttt{\# Compute task-specific $w_t(x_t, y, A)$ according to Eqs~(\ref{eqn:wt_denoise}--\ref{eqn:wt_color})}

    \textcolor{blue}{$x_{i-1} \leftarrow x_{i-1}' - w_t \nabla_{x_t} \left\| y - A \hat{x}_0 \right\|^2$}
}
\KwReturn{$x_0$}
\end{algorithm}

\subsubsection{Denoising}
\label{sec:wt_denoising}
As described in section~\ref{sec:dps-w}, the exact noisy likelihood score for a denoising problem (for which the measurement operator is the identity: $A=I$) is calculated at each time step and the step size $w_t$ chosen to minimize the DPS error,

\begin{equation}
w_t = \arg\min_{w_t} \left\| s_{\theta}(y | x_t) - w_t s_\text{DPS}(y|x_t, I) \right\|^2,
\end{equation}
where $s_\theta(y | x_t)$ and $s_\text{DPS}$ are defined in Eq~(\ref{eqn:ref_noisy_likelihood}) and Eq~(\ref{eqn:s_dps}), respectively. Importantly, the $\zeta_t$ in $s_\text{DPS}$ is set to $1$, so $w_t$ is defined to be a direct replacement for $\zeta_t$ in DPS. The solution is given by the projection of $s_\theta$ onto $s_\text{DPS}$:

\begin{equation}
\label{eqn:wt_denoise}
w_t = \frac{\langle s_\theta(y|x_t),\ s_{\text{DPS}}(y|x_t, I) \rangle}{\left\| s_{\text{DPS}}(y|x_t, I) \right\|^2}.
\end{equation}

\subsubsection{Super resolution}
\label{sec:wt_sr}
The measurement operator for super resolution, $A_\text{SR}$, is the bicubic downsampling matrix \cite{DPS}. Its transpose $A_\text{SR}^T$ performs the corresponding upsampling operation. In this paper we consider $4\text{x}$ super resolution, with scale factor $r=4$.

Denoising of the upsampled measurement, $A_\text{SR}^Ty$, is chosen as a reference task. The step size $w_t$ is fit on this task and transferred to the super resolution problem.

\begin{equation}
\label{eqn:wt_sr}
w_t = r \frac{\langle s_\theta(A_\text{SR}^Ty|x_t),\ s_{\text{DPS}}(A_\text{SR}^Ty|x_t, I) \rangle}{\left\| s_{\text{DPS}}(A_\text{SR}^Ty|x_t, I) \right\|^2}.
\end{equation}

% A multiplicative factor $r$ is introduced to account for the dimensionality of the reference measurement $A_\text{SR}^Ty$ being $r^2$ times greater than that of the down-sampled $y$. Assuming the norm is proportional to $\sqrt{d}$, where $d$ is the dimensionality, we expect the reference residual norm $\|A_\text{SR}^Ty - x_t\|$ to be $~r$ times greater than the residual for the down-sampled image, $\|y - A_\text{SR}x_t\|$. Finally, since DPS is linear in the residual norm and $w_t$ for denoising in Eq~(\ref{eqn:wt_denoise}) has $s_\text{DPS}$ in the numerator and $s_\text{DPS}$ squared in the denominator, we expect the $w_t$ for down-sampled images to be $r$ times greater than up-sampled images.

A multiplicative correction factor $r$ is introduced to account for the larger dimensionality of the reference measurement $A_{\text{SR}}^\top y$, which has $r^2$ times more elements than the downsampled measurement $y$. Assuming the norm of a residual scales proportionally to $\sqrt{d}$, where $d$ is the dimensionality, the residual norm $\|A_{\text{SR}}^\top y - x_t\|$ for the upsampled image is expected to be approximately $r$ times larger than the residual norm for the downsampled input, $\|y - A_{\text{SR}} x_t\|$. Since the DPS score is linear in the residual and the scalar weight $w_t$ (from Eq.~(\ref{eqn:wt_denoise})) is given by a ratio with $s_{\text{DPS}}$ in the numerator and its squared norm in the denominator, $w_t$ scales inversely with the residual norm. Therefore, we expect $w_t$ for downsampled images to be approximately $r$ times larger than the corresponding $w_t$ for upsampled images.

Indeed, for the majority of sampling steps on real super resolution problems, the ratio of the two norms is roughly $4$. However, as $x_t$ becomes a more realistic estimation of $x_0$ towards the end of the sampling process, the reference measurement $y_\text{ref}$ is less suitable and the norm ratio increases sharply. For this reason, we propose to set a parameter $w_\text{max}$ to cap the DPS-w weight. The value $w_\text{max}$ is set to $2.0$ for FFHQ and $3.0$ for ImageNet. For sampling with $100$ steps instead of $1000$, we multiply $w_\text{max}$ by $10$ to account for the increased step sizes. 

\subsubsection{Inpainting}
\label{sec:wt_inpaint}
As discussed in \ref{sec:exact_inpainting}, at dimensions corresponding to unmasked pixels, the inpainting posterior score is well-approximated by the denoising posterior score. This approximation becomes exact when $\sigma_y=0$, despite the intractability of the posterior score in general.

The inpainting measurement operator is the image mask $A_\text{M}=\text{diag}(m_1, \ldots, m_n)$ with $m_i \in \{0, 1\}$. The DPS-w step size $w_t$ is chosen to minimize the error in the unmasked dimensions (i.e. those with $m_i=1$):

\begin{equation}
\label{eqn:wt_inpaint}
w_t = \frac{\langle s_\theta(y|x_t),\ A_\text{M} s_{\text{DPS}}(y|x_t, A_\text{M}) \rangle}{\left\| A_\text{M}s_{\text{DPS}}(y|x_t, A_\text{M}) \right\|^2}.
\end{equation}

For random inpainting with low masking rate, we expect the posterior scores for inpainting and denoising to be similar and $w_t$ to be suitable for masked and unmasked dimensions alike. Recall that the $w_t$ for denoising are shown to have significantly lower magnitude than the $\zeta_t$ of DPS. For higher masking rates, the weights are less likely to be suited for masked dimensions, but in the case of random inpainting, the prior score function is conditioned by local unmasked pixels. For very high masking rates and box or patch inpainting, the reference denoising task is less suitable, and there is reduced influence of the unmasked pixels on the prior score. Although we still see reasonable performance on $95\%$ random inpainting problems, scaling $w_t$ by $\sqrt{d/d_u}$ ($d$ is the total number of dimensions and $d_u$ is the number of unmasked dimensions) improves performance to match DSG results. See \ref{sec:enhanced_guidance} for more details.

\subsubsection{Colorization}
\label{sec:wt_color}
For colorization problems, the measurement operator converts a color image to grayscale by computing a weighted average across the RGB channels, using standard luminance weights $0.2989$, $0.5870$, and $0.1140$ for the red, green, and blue channels, respectively\citep{song2023consistencymodels}. To preserve the original image dimensionality, the resulting grayscale intensities are duplicated across all three channels. Assuming $x_0 \in \mathbb{R}^{3n}$ is a vectorized image with stacked color channels (that is, $x_0 = [x_\text{R};\, x_\text{G};\, x_\text{B}]$, where each $x_c \in \mathbb{R}^n$), the measurement operator $A_\text{col} \in \mathbb{R}^{3n \times 3n}$ is defined as:

\[
A_\text{col} = \begin{bmatrix}
0.2989\, I_n & 0.5870\, I_n & 0.1140\, I_n \\
0.2989\, I_n & 0.5870\, I_n & 0.1140\, I_n \\
0.2989\, I_n & 0.5870\, I_n & 0.1140\, I_n
\end{bmatrix},
\]

where $I_n$ is the $n \times n$ identity matrix.

We choose denoising of the grayscale image as the reference task for DPS-w. The weight $w_t$ is chosen to minimize the error in the DPS guidance with $A=I$, and then applied unchanged to the colorization task.

\begin{equation}
\label{eqn:wt_color}
w_t = \frac{\langle s_\theta(y|x_t),\ s_{\text{DPS}}(y|x_t, I) \rangle}{\left\| s_{\text{DPS}}(y|x_t, I) \right\|^2}
\end{equation}

\section{Further image restoration samples}
\label{sec:further-image}

\begin{figure}[ht]
\centering
\includegraphics[width=\linewidth]{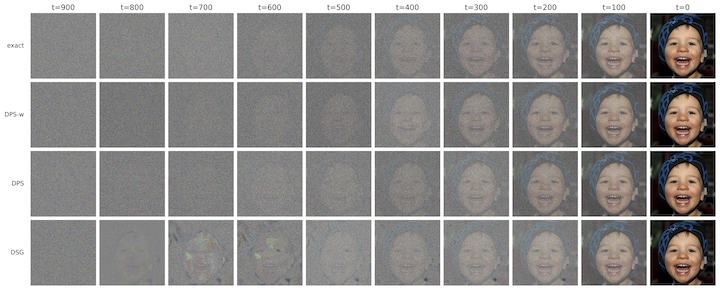}
\caption{Denoising trajectories for an FFHQ image using the exact denoising posterior score equation (in terms approximate prior score $s_\theta$), DPS, DPS-w, and DSG.}
\label{fig:appendix:denoise_trajectory}
\end{figure}

\begin{figure}[ht]
\label{fig:restorations_dsg}
\vskip 0.2in
\begin{center}
\centerline{\includegraphics[width=\columnwidth]{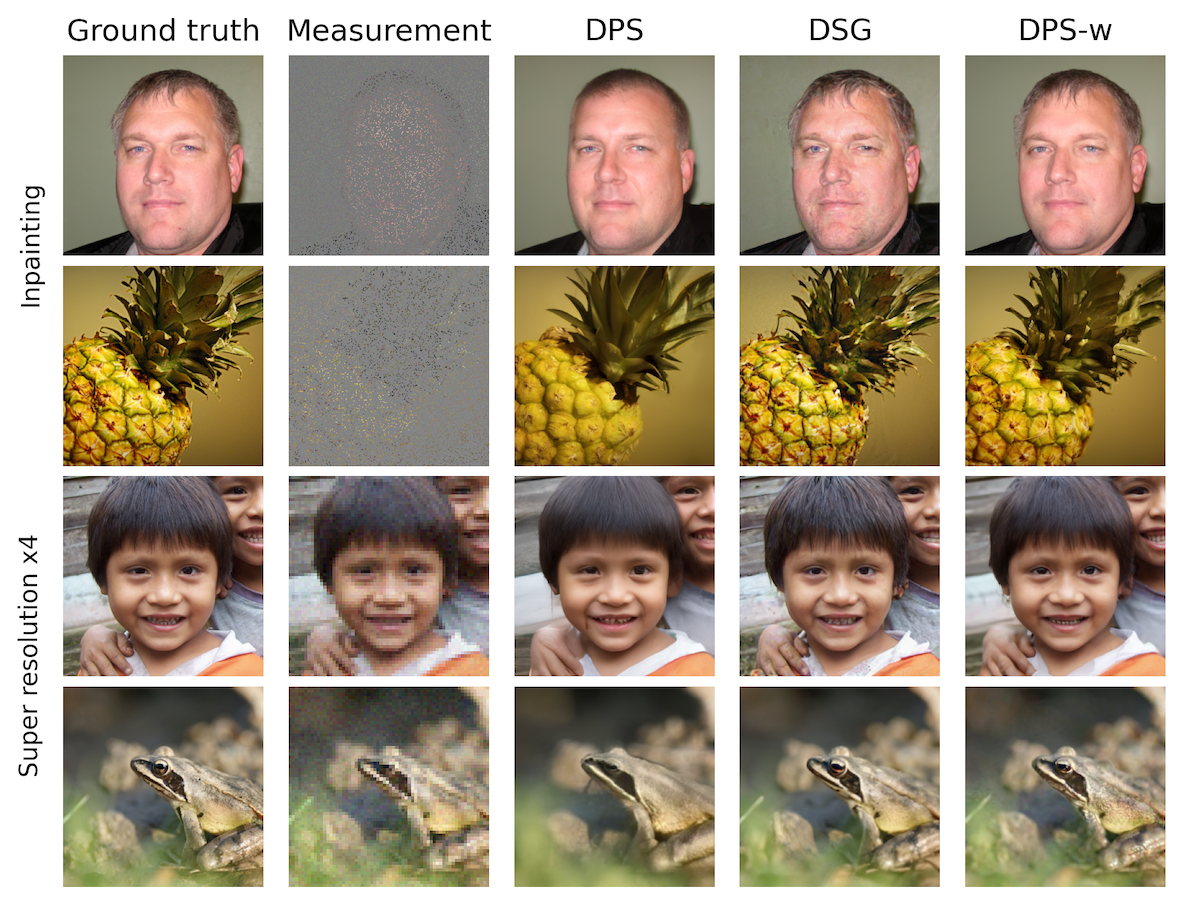}}
\caption{Sample images for inverse linear problems inpainting and super resolution comparing methods DPS, DSG and DPS-w for $\sigma_y = 0.05$.}
\end{center}
\vskip -0.2in
\end{figure}

\begin{figure}[ht]
\vskip 0.2in
\begin{center}
\centerline{\includegraphics[width=\columnwidth]{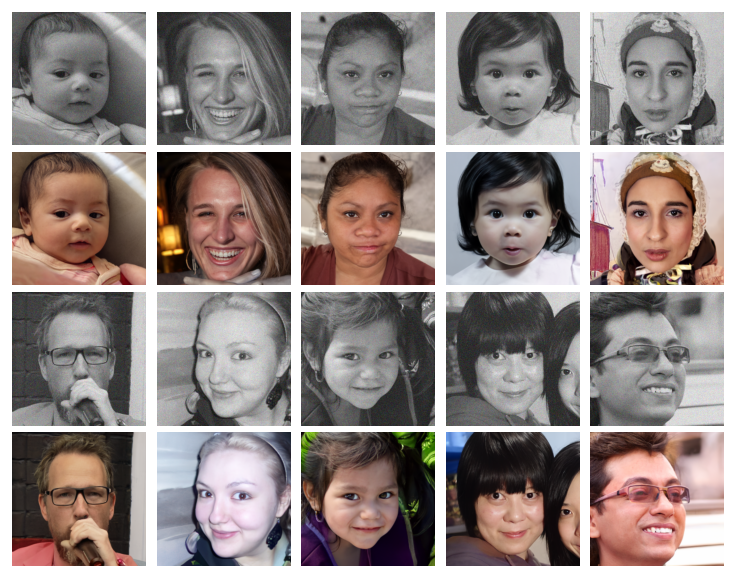}}
\caption{DPS-w samples for the colorization inverse linear problem for $\sigma_y = 0.1$ on FFHQ.}
\label{fig:ffhq_color}
\end{center}
\vskip -0.2in
\end{figure}

\begin{figure}[ht]
\vskip 0.2in
\begin{center}
\centerline{\includegraphics[width=\columnwidth]{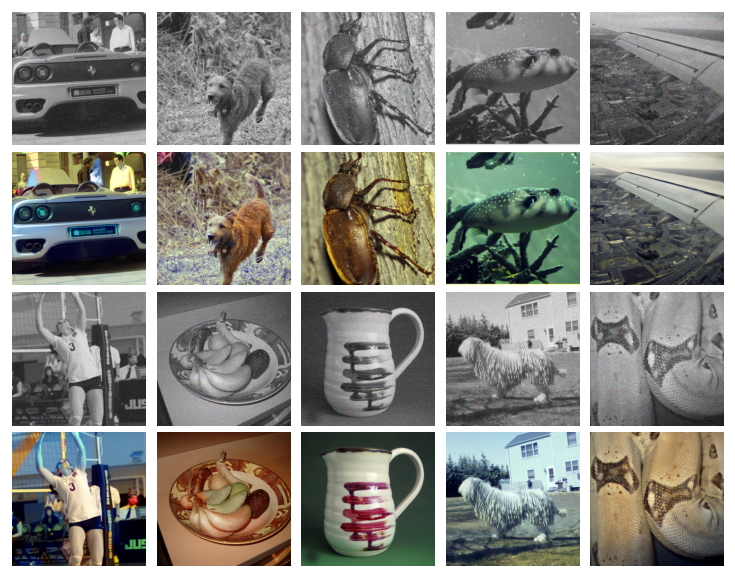}}
\caption{DPS-w samples for the colorization inverse linear problem for $\sigma_y = 0.1$ on ImageNet.}
\label{fig:imagenet_color}
\end{center}
\vskip -0.2in
\end{figure}

%%%%%%%%%%%%%%%%%%%%%%%%%%%%%%%%%%%%%%%%%%%%%%%%%%%%%%%%%%%%%%%%%%%%%%%%%%%%%%%
%%%%%%%%%%%%%%%%%%%%%%%%%%%%%%%%%%%%%%%%%%%%%%%%%%%%%%%%%%%%%%%%%%%%%%%%%%%%%%%
\FloatBarrier
\section{Further quantitative results}
\label{sec:appendix_tables}

\subsection{Necessary conditions of a true posterior sampler}
\label{sec:nec_conds_posterior}
For samples drawn from a true posterior distribution, the expected mean squared error (MSE) is equal to twice the minimum mean squared error (MMSE)\citep{Blau}. Additionally, the residual noise, obtained by subtracting the posterior sample $x_0$ from the observation $y$, should be independent of $x_0$ and normally distributed with variance $\sigma_y^2$, as specified by the measurement model.

Since we introduce a tractable expression for the exact denoising posterior score function in terms of the unconditional (prior) score, we evaluate its correctness by testing whether these two conditions are satisfied on the FFHQ dataset. Because the FFHQ DDPM model and the numerical integration over 1000 discrete time steps are both approximations, we are also assessing the accuracy of the DDPM model itself.

Using Eq~(\ref{eqn:new_form_posterior_VP_resolved_3}), 40 independent samples for each of the first 50 FFHQ images were drawn from the denoising posterior for noise level $\sigma_y=0.2$. For each image, the first sample was used to calculate MSE, while the mean of the remaining 39 samples was used to estimate MMSE. The MSE to MMSE ratio is $1.8$, comparable with the theoretical value of $2.0$ for exact posterior samplers. For reference, the ratio obtained using DPS samples (with $\zeta^\prime=1.0$) is $1.4$.

The residual noise was found to be independent (Pearson correlation coefficient $< 0.01$) and normally distributed with standard deviation of $0.2$ (Kolmogorov-Smirnov test p-value of $0.9$).

\subsection{Enhanced guidance strength for random inpainting}
\label{sec:enhanced_guidance}
As described in Section~\ref{sec:wt_inpaint}, we also conducted inpainting experiments on the ImageNet dataset using an enhanced DPS-w step size strategy. In this variant, the DPS-w step size, $w_t$, is scaled by a factor of $\sqrt{d/d_u}$, where $d$ is the total number of dimensions and $d_u$ is the number of unmasked dimensions. This adjustment increases the strength of the DPS guidance in line with the degree of masking. The enhanced step size leads to improved performance metrics, achieving results that closely match those of DSG. The results are presented in Table~\ref{dps-w-enhanced}.

% \multirow[c]{1}{*}{Task} & \multicolumn{3}{c}{FFHQ} & \multicolumn{3}{c}{ImageNet} \\
% \hspace{1em}Method& LPIPS ($\downarrow$) & SSIM ($\uparrow$) & PSNR ($\uparrow$) & LPIPS ($\downarrow$) & SSIM ($\uparrow$) & PSNR ($\uparrow$) \\
% \midrule
% Inpainting 40\%\\
% \hspace{1em}DAPS & 0.109 & 0.847 & 32.533 & 0.123 & 0.829 & 30.833 \\

\begin{table*}[t]
\caption{Quantitative evaluation for inverse linear problems inpainting and super resolution on FFHQ and ImageNet datasets for $\sigma_y=0.05$. A $*$ indicates that published hyperparameters were available for the particular task. All methods were run with $1000$ DDPM steps, except our DPS-w 100 steps method. See Tables \ref{main_table_ci_ffhq} and \ref{main_table_ci_imagenet} for confidence intervals.}
\label{main_table}
\centering
% \begin{small}
\begin{tabular}{lcccccc}
\toprule
\multirow[c]{1}{*}{Task} & \multicolumn{3}{c}{FFHQ} & \multicolumn{3}{c}{ImageNet} \\
\hspace{1em}Method& LPIPS ($\downarrow$) & SSIM ($\uparrow$) & PSNR ($\uparrow$) & LPIPS ($\downarrow$) & SSIM ($\uparrow$) & PSNR ($\uparrow$) \\
\midrule
Inpainting 40\%\\
\hspace{1em}DAPS & 0.109 & 0.847 & 32.533 & 0.123 & 0.829 & 30.833 \\
\hspace{1em}DPS & 0.143 & 0.865 & 30.301 & 0.233 & 0.778 & 27.553 \\
\hspace{1em}DSG & \textbf{0.051} & 0.930 & 34.139 & \textbf{0.080} & \textbf{0.916} &\textbf{32.721} \\

\hspace{1em}DPS-w (ours) & \underline{0.060} & \underline{0.932} & \underline{34.239} & \underline{0.089} & \underline{0.914} & \underline{32.330} \\
\hspace{1em}DPS-w 100 steps & 0.072 & \textbf{0.936} & \textbf{34.402} & 0.134 & 0.892 & 31.378\\
% \cline{1-7}
\\
Inpainting 70\% \\
\hspace{1em}DAPS$*$ & 0.130 & 0.833 & 30.565 & 0.164 & 0.785 & 28.351 \\
 \hspace{1em}DPS & 0.180 & 0.813 & 27.818 & 0.285 & 0.714 & 25.254 \\
\hspace{1em}DSG & \textbf{0.076} & 0.880 & 30.587 & \textbf{0.108} & \textbf{0.850} & \textbf{28.707} \\
\hspace{1em}DPS-w (ours) & \underline{0.091} & \underline{0.887} & \underline{30.646} & \underline{0.132} & \underline{0.846} & \underline{28.487} \\
\hspace{1em}DPS-w 100 steps & 0.100 & \textbf{0.898} & \textbf{31.060} & 0.187 & 0.827 & 27.974 \\

% \cline{1-8}
\\
Inpainting 92\%\\
\hspace{1em}DAPS & 0.221 & 0.743 & \underline{25.716} & 0.347 & 0.618 & \underline{23.334} \\
\hspace{1em}DPS$*$ & 0.254 & 0.702 & 23.699 & 0.375 & 0.589 & 21.523 \\
\hspace{1em}DSG$*$ & \textbf{0.166} & 0.750 & 25.449 & \textbf{0.233} & 0.646 & 22.979 \\
 % \cdashline{2-8}
\hspace{1em}DPS-w (ours) & \underline{0.177} & \underline{0.773} & 25.601 & \underline{0.261} & \underline{0.669} & 23.225 \\
\hspace{1em}DPS-w 100 steps & 0.180 & \textbf{0.790} & \textbf{25.801} & 0.270 & \textbf{0.690} & \textbf{23.529} \\

\\
Super resolution\\
\hspace{1em}DAPS$*$ & 0.181 & \underline{0.797} & \textbf{28.733} & 0.255 & 0.708 & \textbf{26.272} \\
\hspace{1em}DPS$*$ & 0.224 & 0.723 & 25.175 & 0.348 & 0.600 & 22.711 \\
\hspace{1em}DSG$*$ & \textbf{0.149} & 0.791 & 27.549 & \textbf{0.233} & 0.704 & 25.407 \\
 % \cdashline{2-8}
\hspace{1em}DPS-w (ours) & \underline{0.150} & 0.792 & 27.579 & \underline{0.236} & \underline{0.714} & 25.478 \\
\hspace{1em}DPS-w 100 steps & 0.152 & \textbf{0.809} & \underline{27.980} & 0.238 & \textbf{0.730} & \underline{25.691} \\

\bottomrule
\end{tabular}

% \end{small}
\end{table*}

\begin{table}[ht]
\caption{Quantitative evaluation for inverse linear problems inpainting and super resolution on FFHQ and ImageNet datasets with noise level $\sigma_y=0.01$. None of the benchmark methods were tuned specifically for this noise level. All methods were run with $1000$ DDPM steps.}
\label{table:sigma_0.01}
\centering
\begin{small}
\begin{tabular}{lcccccc}
\toprule
\multirow[c]{1}{*}{Task} & \multicolumn{3}{c}{FFHQ} & \multicolumn{3}{c}{ImageNet} \\
\hspace{1em}Method& LPIPS ($\downarrow$) & SSIM ($\uparrow$) & PSNR ($\uparrow$) & LPIPS ($\downarrow$) & SSIM ($\uparrow$) & PSNR ($\uparrow$) \\
\midrule
Inpainting 40\%\\
\hspace{1em}DAPS & 0.053 & 0.933 & 35.158 & \underline{0.060} & 0.920 & 33.413 \\
\hspace{1em}DPS & 0.116 & 0.895 & 32.008 & 0.209 & 0.803 & 29.002 \\
\hspace{1em}DSG & \underline{0.046} & \underline{0.949} & \underline{35.241} & 0.081 & \underline{0.928} & \underline{33.537} \\
\hspace{1em}DPS-w (ours) & \textbf{0.023} & \textbf{0.966} & \textbf{36.750} & \textbf{0.039} & \textbf{0.951} & \textbf{34.719}\\
\\
Inpainting 70\%\\
\hspace{1em}DAPS & 0.102 & 0.877 & 31.329 & 0.129 & 0.835 & 29.131 \\
\hspace{1em}DPS & 0.145 & 0.853 & 29.447 & 0.251 & 0.747 & 26.501 \\
\hspace{1em}DSG & \underline{0.061} & \underline{0.915} & \textbf{31.792 }& \underline{0.101} & \underline{0.875} & \textbf{29.627}\\
\hspace{1em}DPS-w (ours) & \textbf{0.057} & \textbf{0.917} &\underline{31.772} & \textbf{0.090} & \textbf{0.878} & \underline{29.504} \\

Inpainting 92\%\\
\hspace{1em}DAPS & 0.215 & 0.755 & 25.805 & 0.339 & 0.631 & \underline{23.419} \\
\hspace{1em}DPS & 0.211 & 0.754 & 25.172 & 0.327 & 0.635 & 22.821 \\
\hspace{1em}DSG & \textbf{0.119} & \textbf{0.809} & \textbf{26.590} & \textbf{0.190} & \textbf{0.713} & \textbf{23.900} \\
\hspace{1em}DPS-w & \underline{0.155} & \underline{0.788} & \underline{25.896} & \underline{0.240} & \underline{0.684} & 23.402 \\

Super resolution\\
\hspace{1em}DAPS & 0.178 & 0.805 & \textbf{29.100} & 0.251 & 0.719 & \textbf{26.585} \\
\hspace{1em}DPS & 0.211 & 0.739 & 25.633 & 0.338 & 0.615 & 23.118 \\
\hspace{1em}DSG & \underline{0.131} & \underline{0.812} & 28.209 & \underline{0.223} & \underline{0.727} & 25.833\\
\hspace{1em}DPS-w & \textbf{0.119} & \textbf{0.826} & \underline{28.534} & \textbf{0.205} & \textbf{0.753} & \underline{26.268} \\

\bottomrule
\end{tabular}

\end{small}
\end{table}

\begin{table}[ht]
\caption{Quantitative evaluation for inverse linear problems inpainting and super resolution on FFHQ and ImageNet datasets with noise level $\sigma_y=0.1$. None of the benchmark methods were tuned specifically for this noise level. All methods were run with $1000$ DDPM steps.}
\label{table:sigma_0.1}
\centering
\begin{small}
\begin{tabular}{lcccccc}
\toprule
\multirow[c]{1}{*}{Task} & \multicolumn{3}{c}{FFHQ} & \multicolumn{3}{c}{ImageNet} \\
\hspace{1em}Method& LPIPS ($\downarrow$) & SSIM ($\uparrow$) & PSNR ($\uparrow$) & LPIPS ($\downarrow$) & SSIM ($\uparrow$) & PSNR ($\uparrow$) \\
\midrule
Inpainting 92\%\\
\hspace{1em}DAPS & \underline{0.236} & \underline{0.710} & \textbf{25.449} & 0.377 & \underline{0.562} & \textbf{22.926} \\
\hspace{1em}DPS & 0.289 & 0.659 & 22.397 & 0.419 & 0.545 & 20.460 \\
\hspace{1em}DSG & 0.271 & 0.618 & 23.225 & \underline{0.357} & 0.487 & 21.019 \\
\hspace{1em}DPS-w & \textbf{0.205} & \textbf{0.745} & \underline{24.898} & \textbf{0.292} & \textbf{0.637} & \underline{22.723} \\
\\
Super resolution\\
\hspace{1em}DAPS & 0.312 & 0.703 & \textbf{26.461} & 0.430 & 0.520 & 23.247 \\
\hspace{1em}DPS & 0.244 & 0.698 & 24.250 & 0.364 & 0.578 & 21.997 \\
\hspace{1em}DSG & \underline{0.194} & \underline{0.737} & 26.079 & \textbf{0.279} & \underline{0.641} & \underline{24.043} \\
\hspace{1em}DPS-w & \textbf{0.189} & \textbf{0.754} & \underline{26.358} & \underline{0.289} & \textbf{0.659} & \textbf{24.360} \\

\bottomrule
\end{tabular}

\end{small}
\end{table}

\begin{table*}[ht]
\caption{Quantitative evaluation with 95\% confidence intervals for inverse linear problems inpainting and super resolution on FFHQ dataset for $\sigma_y=0.05$.. A $*$ indicates that published hyperparameters were available for the particular task. All methods were run with $1000$ DDPM steps, except our DPS-w 100 steps method. Confidence intervals are computed as $t s / \sqrt{N}$ assuming normally distributed metrics, where $s$ is the sample standard deviation, $t$ is the critical value from the Student’s t-distribution for $N-1$ degrees of freedom, and $N=200$.}
\label{main_table_ci_ffhq}
\centering
% \begin{small}
\begin{tabular}{lccc}
\toprule
\multirow[c]{1}{*}{Task} & \multicolumn{3}{c}{FFHQ} \\
\hspace{1em}Method& LPIPS ($\downarrow$) & SSIM ($\uparrow$) & PSNR ($\uparrow$)\\
\midrule
Inpainting 40\%\\
\hspace{1em}DAPS & 0.109 $\pm$ 0.003 & 0.847 $\pm$ 0.002 & 32.533 $\pm$ 0.119 \\
\hspace{1em}DPS & 0.143 $\pm$ 0.004 & 0.865 $\pm$ 0.004 & 30.301 $\pm$ 0.225 \\
\hspace{1em}DSG & \textbf{0.051} $\pm$ 0.002 & 0.930 $\pm$ 0.003 & 34.139 $\pm$ 0.197 \\

\hspace{1em}DPS-w (ours) & \underline{0.060} $\pm$ 0.002 & \underline{0.932} $\pm$ 0.002 & \underline{34.239} $\pm$ 0.205 \\
\hspace{1em}DPS-w 100 steps & 0.072 $\pm$ 0.003 & \textbf{0.936} $\pm$ 0.002 & \textbf{34.402} $\pm$ 0.214 \\
% \cline{1-7}
\\
Inpainting 70\% \\
\hspace{1em}DAPS$*$ & 0.130 $\pm$ 0.003 & 0.833 $\pm$ 0.002 & 30.565 $\pm$ 0.199 \\
 \hspace{1em}DPS & 0.180 $\pm$ 0.005 & 0.813 $\pm$ 0.006 & 27.818 $\pm$ 0.216\\
\hspace{1em}DSG & \textbf{0.076} $\pm$ 0.002 & 0.880 $\pm$ 0.005 & 30.587 $\pm$ 0.231\\
\hspace{1em}DPS-w (ours) & \underline{0.091} $\pm$ 0.003 & \underline{0.887} $\pm$ 0.003 & \underline{30.646} $\pm$ 0.248 \\
\hspace{1em}DPS-w 100 steps & 0.100 $\pm$ 0.003 & \textbf{0.898} $\pm$ 0.003 & \textbf{31.060} $\pm$ 0.265\\

% \cline{1-8}
\\
Inpainting 92\%\\
\hspace{1em}DAPS & 0.221 $\pm$ 0.003 & 0.743 $\pm$ 0.005 & \underline{25.716} $\pm$ 0.229\\
\hspace{1em}DPS$*$ & 0.254 $\pm$ 0.006 & 0.702 $\pm$ 0.007 & 23.699 $\pm$ 0.192\\
\hspace{1em}DSG$*$ & \textbf{0.166} $\pm$ 0.004 & 0.750 $\pm$ 0.007 & 25.449 $\pm$ 0.220\\
 % \cdashline{2-8}
\hspace{1em}DPS-w (ours) & \underline{0.177} $\pm$ 0.005 & \underline{0.773} $\pm$ 0.006 & 25.601 $\pm$ 0.249\\
\hspace{1em}DPS-w 100 steps & 0.180 $\pm$ 0.005 & \textbf{0.790} $\pm$ 0.006 & \textbf{25.801} $\pm$ 0.254 \\

\\
Super resolution\\
\hspace{1em}DAPS$*$ & 0.181 $\pm$ 0.003 & \underline{0.797} $\pm$ 0.004 & \textbf{28.733} $\pm$ 0.213 \\
\hspace{1em}DPS$*$ & 0.224 $\pm$ 0.006 & 0.723 $\pm$ 0.007 & 25.175 $\pm$ 0.210 \\
\hspace{1em}DSG$*$ & \textbf{0.149} $\pm$ 0.004 & 0.791 $\pm$ 0.006 & 27.549 $\pm$ 0.213 \\
 % \cdashline{2-8}
\hspace{1em}DPS-w (ours) & \underline{0.150} $\pm$ 0.004 & 0.792 $\pm$ 0.005 & 27.579 $\pm$ 0.222\\
\hspace{1em}DPS-w 100 steps & 0.152 $\pm$ 0.004 & \textbf{0.809} $\pm$ 0.005 & \underline{27.980} $\pm$ 0.214 \\

\bottomrule
\end{tabular}

% \end{small}
\end{table*}

\begin{table*}[ht]
\caption{Quantitative evaluation with 95\% confidence intervals for inverse linear problems inpainting and super resolution on ImageNet dataset for $\sigma_y=0.05$. A $*$ indicates that published hyperparameters were available for the particular task. All methods were run with $1000$ DDPM steps, except our DPS-w 100 steps method. Confidence intervals are computed as $t s / \sqrt{N}$ assuming normally distributed metrics, where $s$ is the sample standard deviation, $t$ is the critical value from the Student’s t-distribution for $N-1$ degrees of freedom, and $N=200$.}
\label{main_table_ci_imagenet}
\centering
% \begin{small}
\begin{tabular}{lccc}
\toprule
\multirow[c]{1}{*}{Task} & \multicolumn{3}{c}{ImageNet} \\
\hspace{1em}Method& LPIPS ($\downarrow$) & SSIM ($\uparrow$) & PSNR ($\uparrow$) \\
\midrule
Inpainting 40\%\\
\hspace{1em}DAPS & 0.123 $\pm$ 0.006 & 0.829 $\pm$ 0.004 & 30.833 $\pm$ 0.282 \\
\hspace{1em}DPS &  0.233 $\pm$ 0.010 & 0.778 $\pm$ 0.016 & 27.553 $\pm$ 0.452 \\
\hspace{1em}DSG & \textbf{0.080} $\pm$ 0.004 & \textbf{0.916} $\pm$ 0.006 &\textbf{32.721} $\pm$ 0.446 \\

\hspace{1em}DPS-w (ours) & \underline{0.089} $\pm$ 0.004 & \underline{0.914} $\pm$ 0.005 & \underline{32.330} $\pm$ 0.401 \\
\hspace{1em}DPS-w 100 steps & 0.134 $\pm$ 0.006 & 0.892 $\pm$ 0.007 & 31.378 $\pm$ 0.383 \\
% \cline{1-7}
\\
Inpainting 70\% \\
\hspace{1em}DAPS$*$ & 0.164 $\pm$ 0.005 & 0.785 $\pm$ 0.008 & 28.351 $\pm$ 0.399 \\
 \hspace{1em}DPS & 0.285 $\pm$ 0.011 & 0.714 $\pm$ 0.018 & 25.254 $\pm$ 0.406 \\
\hspace{1em}DSG & \textbf{0.108} $\pm$ 0.005 & \textbf{0.850} $\pm$ 0.010 & \textbf{28.707} $\pm$ 0.459 \\
\hspace{1em}DPS-w (ours) & \underline{0.132} $\pm$ 0.005 & \underline{0.846} $\pm$ 0.009 & \underline{28.487} $\pm$ 0.440 \\
\hspace{1em}DPS-w 100 steps & 0.187 $\pm$ 0.007 & 0.827 $\pm$ 0.010 & 27.974 $\pm$ 0.432 \\

% \cline{1-8}
\\
Inpainting 92\%\\
\hspace{1em}DAPS & 0.347 $\pm$ 0.009 & 0.618 $\pm$ 0.017 & \underline{23.334} $\pm$ 0.403 \\
\hspace{1em}DPS$*$ & 0.375 $\pm$ 0.011 & 0.589 $\pm$ 0.020 & 21.523 $\pm$ 0.346 \\
\hspace{1em}DSG$*$ & \textbf{0.233} $\pm$ 0.007 & 0.646 $\pm$ 0.017 & 22.979 $\pm$ 0.402 \\
 % \cdashline{2-8}
\hspace{1em}DPS-w (ours) & \underline{0.261} $\pm$ 0.008 & \underline{0.669} $\pm$ 0.017 & 23.225 $\pm$ 0.422 \\
\hspace{1em}DPS-w 100 steps & 0.270 $\pm$ 0.009 & \textbf{0.690} $\pm$ 0.017 & \textbf{23.529} $\pm$ 0.428 \\

\\
Super resolution\\
\hspace{1em}DAPS$*$ & 0.255 $\pm$ 0.007 & 0.708 $\pm$ 0.014 & \textbf{26.272} $\pm$ 0.413 \\
\hspace{1em}DPS$*$ & 0.348 $\pm$ 0.011 & 0.600 $\pm$ 0.020 & 22.711 $\pm$ 0.391 \\
\hspace{1em}DSG$*$ & \textbf{0.233} $\pm$ 0.009 & 0.704 $\pm$ 0.016 & 25.407 $\pm$ 0.407 \\
 % \cdashline{2-8}
\hspace{1em}DPS-w (ours) & \underline{0.236} $\pm$ 0.007 & \underline{0.714} $\pm$ 0.015 & 25.478 $\pm$ 0.392 \\
\hspace{1em}DPS-w 100 steps & 0.238 $\pm$ 0.008 & \textbf{0.730} $\pm$ 0.015 & \underline{25.691} $\pm$ 0.401 \\

\bottomrule
\end{tabular}

% \end{small}
\end{table*}

\begin{table*}[ht]
\caption{Quantitative evaluation of DPS-w with enhanced step size, compared to DPS-w and DSG for random inpainting on ImageNet dataset, $\sigma_y=0.05$. All methods were run with $1000$ DDPM steps.}
\label{dps-w-enhanced}
\centering
% \begin{small}
\begin{tabular}{lccc}
\toprule
\multirow[c]{1}{*}{Task} & \multicolumn{3}{c}{ImageNet} \\
\hspace{1em}Method& LPIPS ($\downarrow$) & SSIM ($\uparrow$) & PSNR ($\uparrow$) \\
\midrule
Inpainting 40\%\\
\hspace{1em}DSG & \underline{0.080} & \underline{0.916} &\underline{32.721} \\

\hspace{1em}DPS-w & 0.089 & 0.914 & 32.330 \\
\hspace{1em}DPS-w enhanced & \textbf{0.072} & \textbf{0.922} & \textbf{32.865} \\
% \cline{1-7}
\\
Inpainting 70\% \\
\hspace{1em}DSG & \textbf{0.108} & \textbf{0.850} & \textbf{28.707} \\
\hspace{1em}DPS-w & \underline{0.132} & 0.846 & 28.487 \\
\hspace{1em}DPS-w enhanced & \textbf{0.108} & \underline{0.847} & \underline{28.608} \\

% \cline{1-8}
\\
Inpainting 92\%\\
\hspace{1em}DSG & \textbf{0.233} & \underline{0.646} & \underline{22.979} \\
 % \cdashline{2-8}
\hspace{1em}DPS-w & 0.261 & \textbf{0.669} & \textbf{23.225} \\
\hspace{1em}DPS-w enhanced & 0.239 & 0.640 & 22.973 \\

\bottomrule
\end{tabular}

% \end{small}
\end{table*}

\FloatBarrier
\section{Technical figures}
\label{sec:tech_figures}

\begin{figure}[ht]
\centering
\includegraphics[width=0.8\columnwidth]{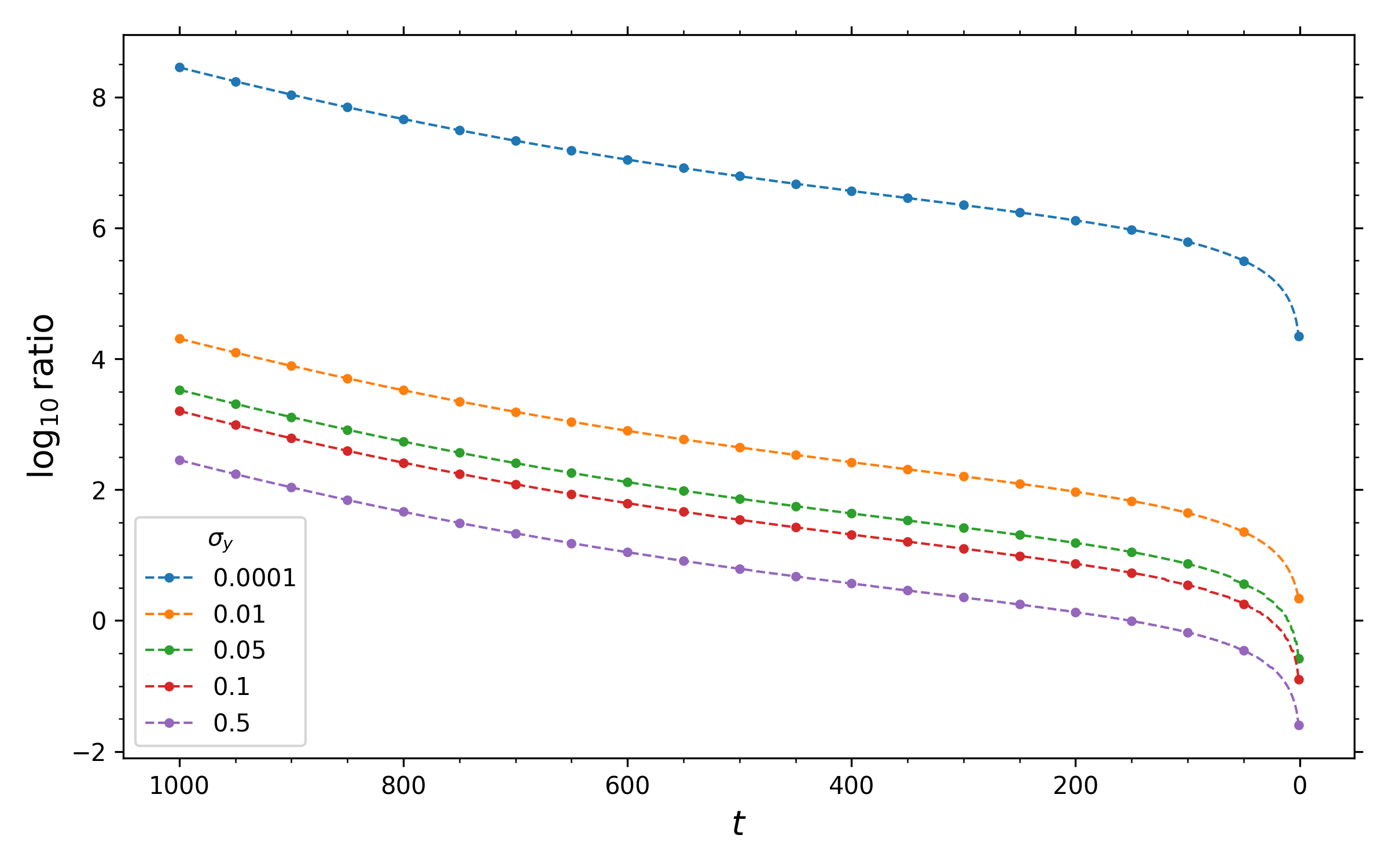}
\caption{The relative strengths of the prior score and the linear guidance term in the denoising posterior score, Eq (\ref{eqn:denoising_score}), for different noise levels $\sigma_y$. The standard logarithm of the guidance norm divided by the prior score norm is plotted against $t$ for denoising an FFHQ image. The linear guidance term is dominant, especially for low $\sigma_y$, but the prior score term has influence towards the end of the process for moderate $\sigma_y$.}
\label{fig:term_ratio}
\end{figure}

\begin{figure}[ht]
\centering
\includegraphics[width=0.8\linewidth]{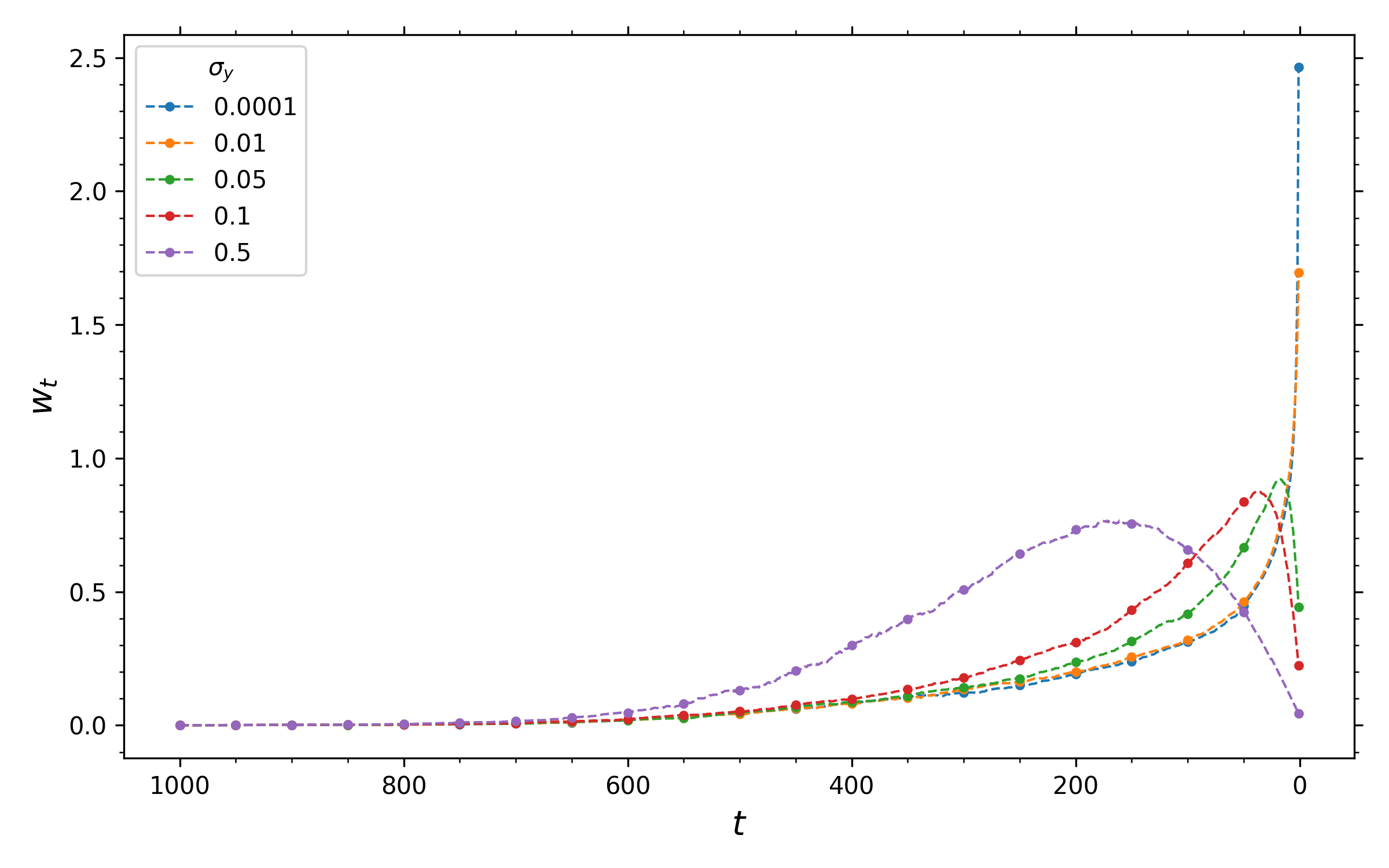}
\caption{Time-dependent $w_t$ calculated as optimum corrective weight for DPS with $\zeta^\prime = 1$; the plotted values can be interpreted as direct alternatives to $\zeta^\prime$ at each $t$. The DPS guidance strength for denoising is far too large for the majority of time steps -- for all $\sigma_y$, the optimum weights are very small for the first half of the trajectory. For low $\sigma_y$, the weights increase monotonically, rising sharply for the last few time steps; for higher $\sigma_y$, the weights peak and begin to fall towards the end of the trajectory.}
\label{fig:wt_plot}
\end{figure}

\begin{figure}[ht]
\centering
\includegraphics[width=\columnwidth]{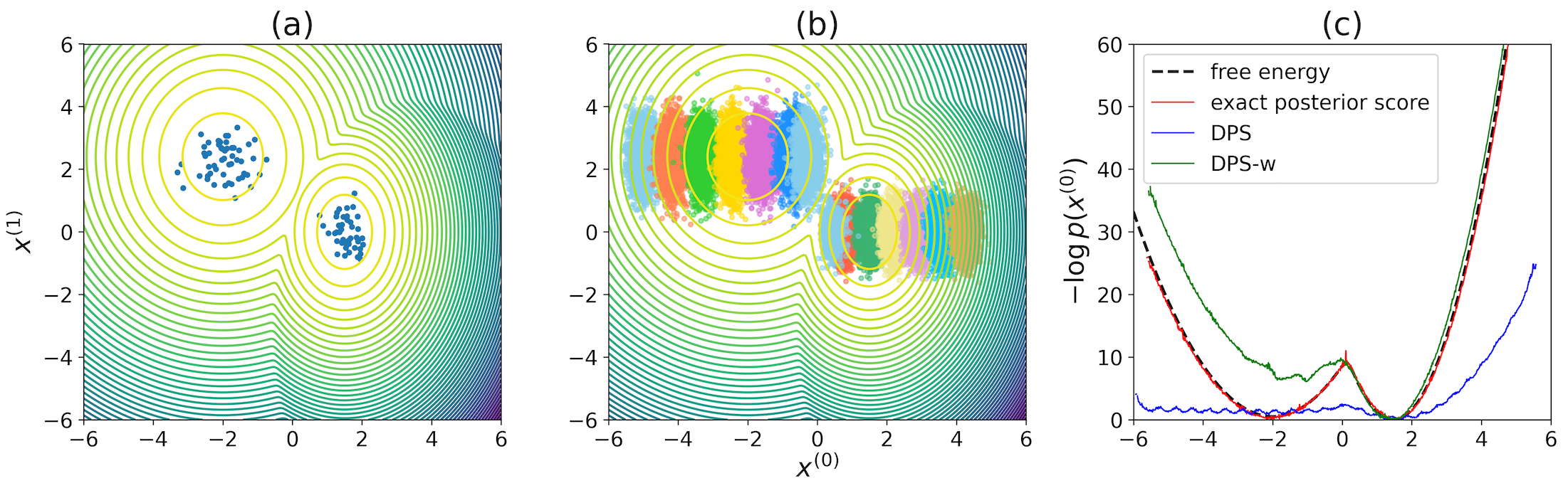}
\caption{Contour plots of a toy Double Well in 2D with (a) 100 unconditional samples and (b) 3000 samples drawn using the exact posterior score function at each Umbrella Sampling window along the $x^{(0)}$ dimension, equivalent to noisy inpainting; (c) the ground truth free energy profile and those computed with samples from different posterior score functions.}
\label{fig:DW}
\end{figure}

\FloatBarrier

\section{Further experimental details}
\subsection{Compute resources}
All experiments were run on single NVIDIA A100 GPUs with 80GB of memory. Time of execution for drawing $200$ samples for an inverse problem for one benchmark method was roughly $4$ hours for FFHQ and $9$ hours for ImageNet. Total compute resources for all experiments was approximately $600$ GPU hours. The additional compute resources allocated to experiments not included in the paper is negligible.

\subsection{Use of existing assets}
The FFHQ\citep{karras2019style} dataset is released by NVIDIA Corporation under Creative Commons BY-NC-SA 4.0. See \url{https://github.com/NVlabs/ffhq-dataset/blob/master/LICENSE.txt} for full license. In our experiments, we used the first $200$ images of the FFHQ validation set, as used in DPS\citep{DPS}. That is, images with label numbers $0$--$199$ were used.

The ImageNet\citep{deng2009imagenet} dataset was in our experiments was version ILSVRC2012. Terms and conditions of use detailed at \url{https://image-net.org/download}. A random sample of $200$ images was taken from the validation set. The unique image ID numbers of the 200 sample images are: 00000022, 00000042, 00000606, 00000906, 00000913, 00000992, 00001493, 00001759, 00001767, 00001958, 00002415, 00004417, 00004546, 00004561, 00004658, 00005437, 00006398, 00007310, 00007441, 00007592, 00007814, 00008001, 00008352, 00008809, 00008813, 00008949, 00009073, 00009533, 00010126, 00010266, 00010346, 00010398, 00010541, 00010949, 00011303, 00011311, 00011411, 00011846, 00012384, 00012440, 00012516, 00012582, 00012685, 00012890, 00013390, 00014493, 00014543, 00015129, 00015311, 00015366, 00016171, 00016468, 00016700, 00017092, 00017680, 00017727, 00017934, 00018069, 00018170, 00018230, 00018305, 00018311, 00018571, 00018702, 00018911, 00019039, 00019232, 00019302, 00019403, 00019627, 00019957, 00020179, 00020320, 00020495, 00020837, 00021052, 00021659, 00021843, 00021882, 00022188, 00022329, 00022399, 00022485, 00022487, 00022628, 00022699, 00023014, 00023027, 00023461, 00023655, 00023857, 00024431, 00024458, 00024472, 00024556, 00025049, 00025405, 00026198, 00026536, 00026638, 00026663, 00027227, 00027468, 00027470, 00027743, 00027766, 00027956, 00028232, 00028524, 00028540, 00028749, 00028812, 00028820, 00029086, 00029318, 00029401, 00029684, 00029692, 00029968, 00031123, 00031622, 00031821, 00032034, 00032279, 00032360, 00032689, 00032798, 00032935, 00032995, 00033482, 00033502, 00033574, 00034010, 00034211, 00034471, 00034687, 00034871, 00035020, 00035269, 00035516, 00036694, 00037323, 00037353, 00037427, 00037569, 00037648, 00038052, 00038875, 00039135, 00039194, 00039517, 00039702, 00039853, 00040215, 00040261, 00040684, 00040724, 00041106, 00041706, 00041815, 00041958, 00041993, 00042291, 00042656, 00043257, 00043291, 00043479, 00043877, 00043926, 00044276, 00044842, 00044964, 00045217, 00045229, 00045300, 00045485, 00045875, 00045912, 00046066, 00046175, 00046236, 00046256, 00046273, 00046486, 00046622, 00046740, 00046743, 00046910, 00047001, 00047290, 00047292, 00047869, 00048116, 00048271, 00048674, 00049260, 00049311, 00049325, 00049509, 00049765.

The benchmark methods DPS\citep{DPS}, DAPS\citep{DAPS}, and DSG\citep{DSG} have code available at the following URLs: \url{https://github.com/DPS2022/diffusion-posterior-sampling}, \url{https://github.com/zhangbingliang2019/DAPS}, and \url{https://github.com/LingxiaoYang2023/DSG2024}, respectively. The method implementations in these repositories were used to run benchmarking experiments. DPS-w was implemented on top of the DPS code base.

%%%%%%%%%%%%%%%%%%%%%%%%%%%%%%%%%%%%%%%%%%%%%%%%%%%%%%%%%%%%

\end{document}